# Higher-order Knowledge Transfer for Dynamic Community Detection with Great Changes

Huixin Ma, Kai Wu, *Member*, *IEEE*, Handing Wang, *Member*, *IEEE*, Jing Liu, *Senior Member*, *IEEE*

***Abstract*—Network structure evolves with time in the real world, and the discovery of changing communities in dynamic networks is an important research topic that poses challenging tasks. Most existing methods assume that no significant change in the network occurs; namely, the difference between adjacent snapshots is slight. However, great change exists in the real world usually. The great change in the network will result in the community detection algorithms are difficulty obtaining valuable information from the previous snapshot, leading to negative transfer for the next time steps. This paper focuses on dynamic community detection with substantial changes by integrating higher-order knowledge from the previous snapshots to aid the subsequent snapshots. Moreover, to improve search efficiency, a higher-order knowledge transfer strategy is designed to determine first-order and higher-order knowledge by detecting the similarity of the adjacency matrix of snapshots. In this way, our proposal can better keep the advantages of previous community detection results and transfer them to the next task. We conduct the experiments on four real-world networks, including the networks with great or minor changes. Experimental results in the low-similarity datasets demonstrate that higher-order knowledge is more valuable than first-order knowledge when the network changes significantly and keeps the advantage even if handling the high-similarity datasets. Our proposal can also guide other dynamic optimization problems with great changes.**

## I. Introduction

Community detection plays a crucial role in representing many real-world complex systems [1], [2], [3]. In finding out the change of networks over time, researchers introduce a concept, dynamic community detection (DCD). Dynamic communities capture the network structures varying with time. Dynamic interaction graphs are applied ubiquity in many fields, such as social networks [4] and protein networks [5].

For this reason, there are many studies aimed at finding and studying this structure, such as spectral clustering [6], [7], deep learning [8], [45], [54], and nonnegative matrix factorization [53]. In general, the DCD problem is NP-hard, and then many efforts have employed evolutionary algorithms (EAs) to solve this problem effectively, and EAs have become one of the leading solutions [10], [11]. How to promote the performance of EAs on this problem is also the main focus of this paper. Aynaud *et al.* [9] proposed a static community detection algorithm for evolving networks. A pattern of community detection with temporal smoothness is formulated as a multi-objective problem [10], [11]. A diffusion model is presented in [12] for evolving networks. Evolutionary-based clustering approaches are introduced to maximize cluster accuracy and minimize the temporal cost from one time step to the successive one [10], [13-15]. However, in the real world, the communities may keep their topological features altered dramatically in the process of evolution, and some types of networks with changes after perturbing are introduced [20]. Great changes occur when nodes and edges are removed, added, and changed in the network beyond a certain threshold. Although existing algorithms have demonstrated superior performance in dynamic networks with minor changes, there is no guarantee that those algorithms still have reasonably good performance in networks with great changes.

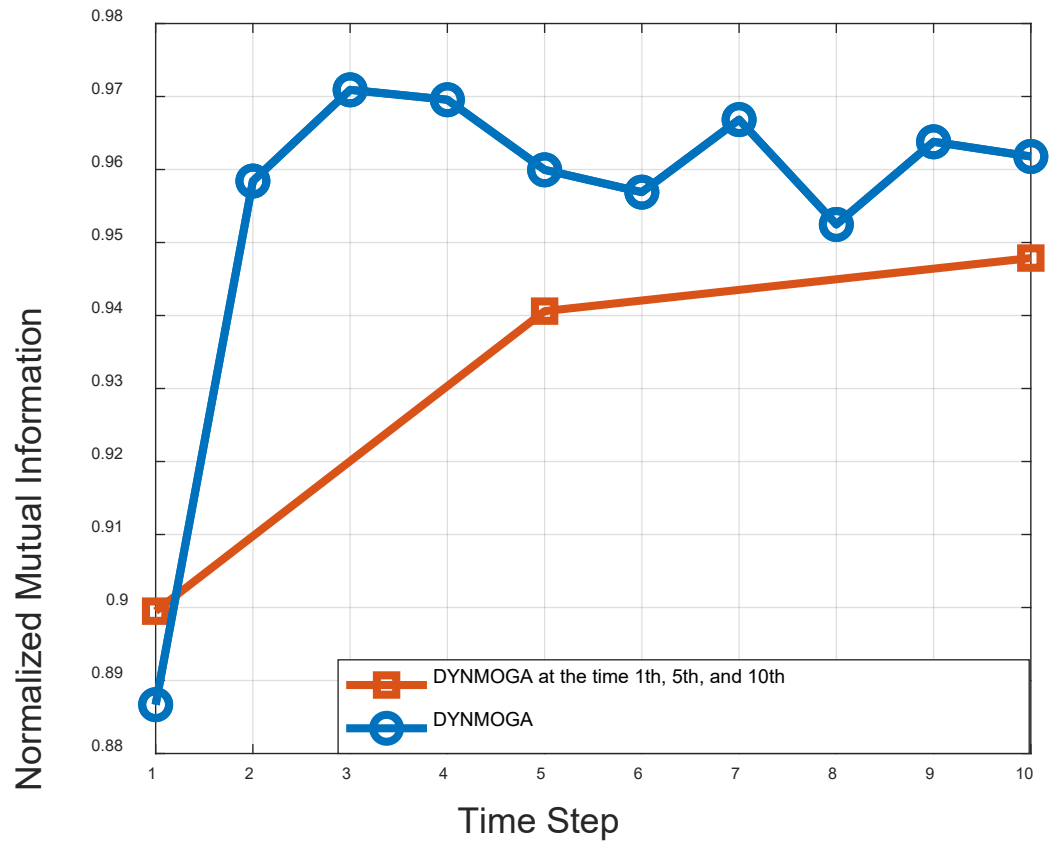

Figure 1 Influence of great changes in community detection. The SYN-FIX dataset's 1st, 5th, and 10th time steps are selected to run DYNMOGA [10] to introduce great changes.

To show the impact of the great network changes on the performance of algorithms, we run the representative DYNMOGA [10] algorithm at the 1$^{st}$, 5$^{th}$, and 10$^{th}$ time steps in the SYN-FIX dataset. As shown in Fig. 1, DYNMOGA has greater accuracy in the successive time steps than in the intermittent time step. It reveals that DYNMOGA may neglect

This work was supported in part by the Key Project of Science and Technology Innovation 2030 supported by the Ministry of Science and Technology of China under Grant 2018AAA0101302, in part by the National Natural Science Foundation of China under Grant 62206205, and in part by the Zhejiang Lab's International Talent Fund for Young Professionals.

Huixin Ma, Kai Wu, and Handing Wang are with the School of Artificial Intelligence, Xidian University, Xi'an 710071, China (e-mail: 948181384@qq.com kwu@xidian.edu.cn, hdwang@xidian.edu.cn).

Jing Liu is with the Guangzhou Institute of Technology, Xidian University, Guangzhou 510555, China (e-mail: neouma@mail.xidian.edu.cn).

the negative effect of the great changes, and it is of great significance that pays attention to this crucial point. It is believed that vertexes and edges do not change randomly in those snapshots. Generally, there is a hypothesis that the deviation from one time step to a successive one is negligible in most state-of-the-art algorithms. The performance of those algorithms decreases when the fact goes against the assumption.

These two characteristics inspire our solution: 1) on the one hand, it is feasible to utilize those unchanged features to improve the performance of the dynamic community detection; 2) on the other hand, the existing dynamic community detection algorithms neglect the network with great changes, which leads to a negative transfer from the previous snapshot to the following snapshot. However, current methods only consider the first characteristic and ignore the second one, as shown in Fig.1. Therefore, there is an urgent need to design a strategy to balance the impact of negative transfer and positive transfer.

This paper proposes a higher-order knowledge transfer for a multi-objective genetic algorithm named HoKT (higher-order knowledge transfer), integrating the designed higher-order normalized mutual information (HoNMI) as one of the objective functions to overcome the shortcoming that the great changes can result in a negative transfer. Many dynamic community detection algorithms [10], [21], [22], [23] introduce a concept of temporal smoothness to minimize the differences between the community structure at the current time step and that obtained at the previous time step. However, after the severe disturbance of the network, the temporal smoothness hurts the performance of the algorithms. HoKT mainly overcomes this adverse effect by introducing the concept of higher-order temporal smoothness. We measure the overlapping degree between the current and previous snapshots when the network changes. This operation can not only reduce the weight of the first-order temporal smoothness but also utilize the higher-order knowledge.

At first, we conduct experiments to confirm that low similarity exists. Secondly, the experiments on synthetic and real-life networks show that HoKT performs better, including the convergence rate, accuracy, and efficiency, than the existing algorithms in solving the dynamic community problem with minor changes. Moreover, we transform four datasets, SYNFIX, Enron, Cell Phone Call, and Four Events, as networks with great changes. The experimental results show that HoKT has an excellent performance in the network with great changes and can balance the impact of negative transfer and positive transfer.

The main contributions of HoKT are summarized as follows:

1) Great changes often occur in dynamic networks. A large amount of negative transfer degrades the current community detection algorithms' performance because they assume that adjacent snapshots have high similarity. This paper is the first work to focus on the dynamic community detection problems with great changes and then give a simple but effective solution.
2) We propose the higher-order knowledge transfer strategy to reduce the negative effect but enhance the positive effect when great changes occur by measuring the overlapping degree of high-order temporal smoothness between two adjacent snapshots of communities.
3) The proposed strategy also has a good reference for other dynamic optimization problems with great changes. We could consider how to utilize the results of early timestamps to achieve higher-order knowledge transfer so that it balances the effect of negative and positive transfer.

The rest of this paper is organized as follows. Backgrounds of the dynamic community detection and multi-objective algorithm are presented in Section II. Section III gives a review of dynamic community detection. Section IV describes the proposed approach, the HoKT, including the objective functions and operators used. Experimental results are reported in Section V. Section VI concludes our proposal and discusses future work.

## II. Related Work

Our proposal mainly focuses on transferring helpful knowledge of the previous tasks to promote the performance of the following tasks. This section introduces the current dynamic community detection methods. Most of them can be classified as exploiting transfer learning to solve the problem of dynamic multi-objective community detection. Moreover, we introduce the basic definitions of multi-objective optimization. It is inefficient to detect the current dynamic community using previous knowledge when there is a low similarity between the current and previous networks. This paper aims to reduce the negative transfer effect when great changes occur. We also show the studies of transfer learning for dynamic multi-objective optimization problems (DMOPs).

### A. Multi-objective Optimization

Since most of the current evolutionary-based dynamic community detection algorithms model the dynamic community detection problem as a multi-objective optimization problem (MOP), and our proposal is also multi-objective optimization, this section introduces the basic knowledge of multi-objective optimization problems. Evolutionary multi-objective optimization, whose primary goal is to deal with multi-objective optimization problems, has attracted increasing interest in the evolutionary computation community [48]. Many scholars apply the non-dominated individuals to multi-objective optimization problems, for example, the non-dominated sorting-based multi-objective evolutionary algorithm (NSGA) [49].

Generally, a MOP with $n$ decision variables and $m$ objective variables can be formulated as follows:

$$\begin{aligned} \min \quad & y = F(x) = (f_1(x), f_2(x), \ldots, f_m(x))^{\mathrm{T}} \\ \textbf{\textit{s.t.}} \quad & g_i(x) \geq 0, i = 1, 2, \ldots, q \\ & h_j(x) = 0, j = 1, 2, \ldots, p \end{aligned} \tag{1}$$

where $x = (x_1, x_2, \ldots, x_n) \subseteq X \in R^n$ is a vector of design variables, and $n$ is the number of independent variables $x_i$. $y = (y_1, y_2, \ldots, y_m) \subseteq Y \in R^m$ is a vector of objective functions. $Y$ is the target space. $F(x)$ maps the solution space into the objective function space. $g_i(x) \geq 0$, $i$=1, 2, …, $q$ is $q$ inequality constraint, and $h_i(x)$=0, $j$=1, 2, ..., $p$ is equality constraint. $F(x)$ is a vector of competing objectives in real-life problems and must be optimized simultaneously. Hence, it is difficult to find a unique solution to the problem. A reasonable solution to a multi-objective problem is to investigate a set of solutions using the Pareto optimality theory [50].

A feasible solution $x_1$ is said to dominate another feasible solution $x_2$ ($x_1 \succ x_2$), if and only if, $f_k(x_1) \le f_k(x_2)$ for $k = 1, 2, ..., m$. A solution is called Pareto optimal if it is not dominated by any other solution in the solution space. A Pareto optimal solution cannot be improved with respect to any objective without worsening at least one other objective. The Pareto optimal set (PS) consists of all optimal solutions in $X$. For a given Pareto optimal set, the corresponding objective function values in the objective space are said to be the Pareto front (PF). In recent years, most real-life problems have been equipped with multi-objective optimization. In this paper, the multi-objective approach cooperating with transfer learning tries to detect communities in dynamic networks.

### B. Dynamic Community Detection Methods

Individuals in the networks interact in diverse ways. A central issue in studying societies is identifying communities [2]. Membership in networks frequently changes, thus significantly characterizing a community structure in dynamic networks. Several ways have been proposed to detect communities in dynamic networks [38-41]. Several methods for detecting communities in dynamic networks offered before 2010 were described in [42]. Convenient methods for detecting communities in dynamic networks are to view the network as small snapshots based on the time stamps and employ static algorithms directly on each network snapshot [38], [43]. However, using static algorithms repeatedly is computationally expensive, especially when the number of snapshots is excellent. Accordingly, another way to detect the communities is using the feature of the network varieties.

Inspired by the evolutionary clustering framework, Folino *et al*. [10] proposed dynamic multi-objective genetic algorithms (DYNMOGA) to detect community structure in dynamic networks. The first objective is to maximize the modular structures at the current time step, and the second objective aims to minimize the divergence between the community structure at the current time step and that obtained at the previous time step. Zeng *et al*. [16] introduced the concept of consensus community to avoid the preceding issue of DYNMOGA. In this way, they guide the current population toward a direction similar to the community structure at the previous time step. Niu *et al*. [17] employed the label propagation method to initialize the communities and restrict the conditions of the mutation process of genetic algorithms, which further improved the detection efficiency and effectiveness. Liu *et al*. [21] discovered some drawbacks with respect to dynamic community detection. Accordingly, they developed a migration operator cooperating with the classic genetic operators to search for inter-community connections. To overcome the drawbacks of community detection, such as the absence of error correction and the NP-hard of modularity-based community detection, Yin *et al*. [44] proposed an effective multi-objective particle swarm optimization to handle dynamic community detection (DYNMODPSO). To consider the nonlinear characteristics of networks, Wang *et al*. [45] proposed a semi-supervised algorithm to overcome the effects of nonlinear properties on low-dimensional representation. Otherwise, a few ways to tackle the problem of discovering dynamic communities in weighted graphs is proposed in [46], [56]. Huang et al. [47] investigate a compact and elegant community pattern based on the k-truss concept, which has advantages in computational costs and efficiency over previous algorithms to update the communities in both large and dynamic networks.

Although significant efforts have been made to discover dynamic communities, there are still drawbacks. The algorithms assume that the abrupt change in the network does not occur. The above algorithms either do not integrate the temporal smoothness into the dynamic community detection problem or consider that the network structures of two adjacent time steps always tend to be similar in the transfer learning process. None of the current algorithms consider the network changes' negative effect. In order to eliminate the negative impact, our algorithm automatically selects high-order temporal smoothness or low-order temporal smoothness to detect dynamic communities. First-order temporal smoothness plays an important role when the network structure does not change abruptly; in contrast, high-order temporal smoothness reduces the negative effect of temporal smoothness when the network changes significantly.

### C. Transfer Learning for DMOPs

In recent years, many advancements have been made in solving dynamic multi-objective optimization problems based on transfer learning [24]. Jiang *et al*. [25] proposed an algorithmic framework that exploits the transfer learning technique as a tool to solve dynamic multi-objective optimization problems (DMOPs). The problem is that low-quality individuals exist during migration, so obtaining substantial improvements in conventional transfer learning is difficult. To overcome the difficulty of the negative transfer, an imbalanced transfer learning method KT-DMOEA [26], is proposed by adjusting the weights of the solutions and employing the knee points. Jiang *et al*. [27] implemented a novel memory-driven manifold transfer learning-based evolutionary algorithm for dynamic multi-objective optimization, which combines the technique of memory to remain the best individuals from the past with the feature of manifold transfer learning to predict the optimal individuals in the new time step in the process of the evolution. Liu *et al*. [28] employed the framework of transfer learning-assisted multi-objective evolutionary clustering with decomposition to handle high-dimensional datasets. In 2021, Zhang *et al*. [29] combined the centroid distance into NSGA-III with transfer learning for DMOPs, named TC_NSGAIII. Liu *et al*. [30] have characterized a transfer learning algorithm that retains the prediction method considering sufficient historical information and modifying the solutions provided by the population prediction strategy to construct the initial population for optimization in the new environment. Two novel operations (clustering-based transfer learning method and clustering-based transfer) are suggested in [31] to reduce negative transfer and find more effective transferred solutions. Jiang *et al.* [32] proposed an individual transfer-based dynamic multi-objective evolutionary algorithm, which filters out some high-quality individuals from historical optimal solutions to avoid negative transfer. Some approaches aim to overcome the obstacle of low accuracy in the linear correlation instances and insufficient training samples. For example, Xu *et al.* [33] proposed an incremental support vector machine-based dynamic multi-objective evolutionary algorithm. The dynamic multi-objective optimization problems are treated as an online learning process, updating the support vector machine with the historical optimal solution. Considering the change of a dynamic multi-objective optimization problem that may occur in both decision and objective spaces, Zhou *et al.* [34] proposed an evolutionary search algorithm with multi-view prediction that uses a kernelized autoencoding model in a reproducing Hilbert space to obtain the multi-view prediction. Feng *et al.* [32] proposed a prediction strategy by extending the autoencoding evolutionary search for solving the dynamic multi-objective problem.

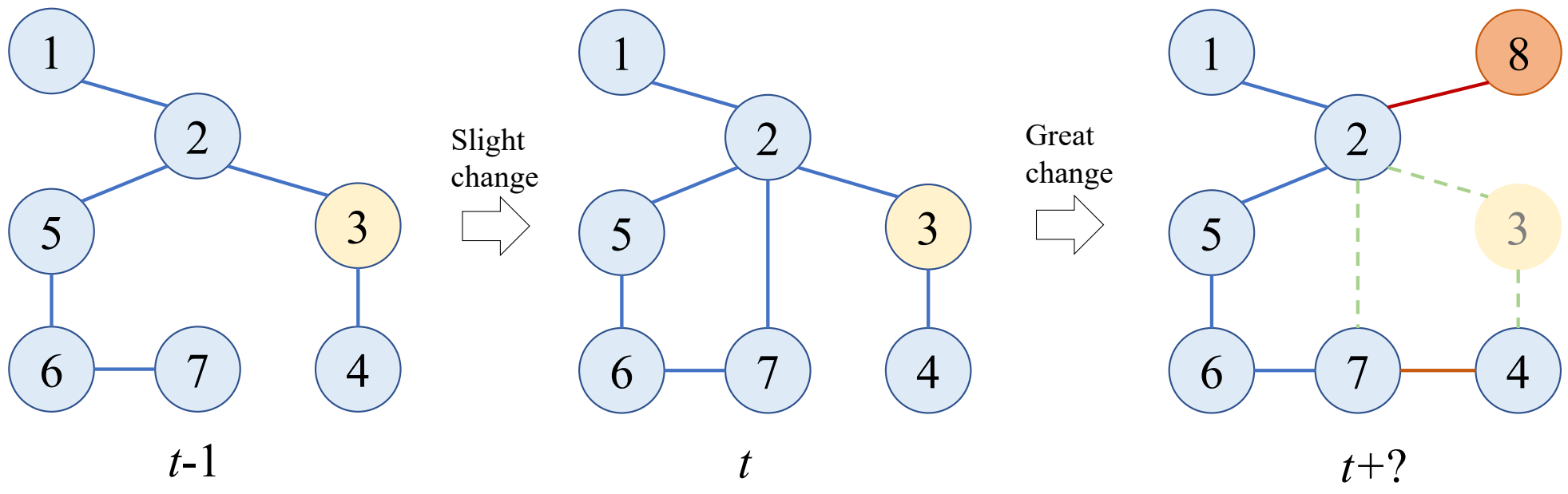

Figure 2 Construction of a multitasking AUC optimization environment.Fig.2. The example of the dynamic network changes. **Slight changes** between time $t$-1 and time $t$: the edge (2, 7) is added. **Great changes** between time $t$ and time $t$+?: Node 3 is deleted, and node eight is added. The edges (2, 7), (2, 3), (3, 4) are deleted, and the edges (2, 8), (4, 7) are added.

Most of the current transfer learning-based dynamic multi-objective evolutionary algorithms are applied to handle continuous optimization problems, such as [27], [35], [36]. Few of those algorithms focus on solving the discrete optimization problem, i.e., they cannot be applied to detect dynamic communities. To integrate transfer learning into dynamic community detection, HoKT introduces the concept of higher-order temporal smoothness, using the previous knowledge indirectly. Moreover, our strategy can also guide other dynamic optimization problems with great changes

## III. HoKT

This section introduces the proposed HoKT. We first introduce the newly designed objective functions used in this paper. Second, we describe the designed higher-order knowledge transfer strategy and then present the essential parts of MOGA. Finally, the overall procedure of HoKT is elaborated.

### *A. Dynamic Community Detection with Great Changes*

A static network is modeled as a graph $G$=($V$, $E$), where $V$ is a set of nodes and $E$ is a set of edges, which connect two elements of $V$. A dynamic network is a sequence $G$={$G^1$, $G^2$, ..., $G^T$}, where each $G^t$ is a snapshot of nodes and connections among these nodes at time $t$, and $t$={1, 2, ..., $T$} is a finite set of time steps. The difference between $G^t$ and $G^{t-1}$ is that some nodes and edges are removed, or new nodes and edges could be added. The dynamic network changes are shown in Fig. 2. We introduce the following two cases: 1) slight network change and 2) great network change. Slight network change is a problem considered by most of the current methods. For example, in Fig. 2, there is only a slight change between time steps $t$-1 and $t$. Great network change is a new scenario proposed in this paper. The change between the two steps is dramatic. For example, in Fig. 2, the change between time steps $t$ and $t$+? is that node three is deleted, but node eight is added. Furthermore, the edges (2, 3), (2, 4), (3, 4), and (2, 7) are removed, and the edges (5, 8) and (2, 8) are appended. The community structure at time step $t$ can be formulated as $CR^t$={$C_1^t$, $C_2^t$, ..., $C_k^t$}, where $CR^t$ is the community structure set at time step t, and each $C_i^t$ in $CR^t$ is a portion of $G^t$ composed of nodes in $G^t$. For each node $V_i^t \in C_i^t \in CR^t$ and $V_j^t \in C_j^t \in CR^t$ , the term $V_i^t \cap V_j^t = \varnothing$ is satisfied.

In general, in terms of overlapping ratio, if the difference between the adjacent snapshots is more significant than 0.01, we can call it a great change. At this time, the influence of negative transfer may become stronger. This is not a strict definition; we are just describing a phenomenon. Moreover, we find that the difference between the adjacent snapshots is less than 0.005 for current datasets (see Figs. 8-14).

### *B. New Objective Functions*

The objective functions we consider are modularity [51] and the proposed HoNMI. Modularity is widely known as a criterion for evaluating the equality of community structure. A community with high modularity has a high link density among

**Algorithm 1:** ***the procedure of computing the ratio***

**Input**:
$t$: time step;
W_Cube{$t$}: the adjacency matrix at time step $t$;
W_Cube{$t$-1}: the adjacency matrix at time step $t$-1;
**Output**:
**Result**: the ratio $r$;

```
t←∅;
for t=2 to T+ do
  for i=1 to n do
    for j=1 to n do
      if (W_Cube{t}(i,j) == W_Cube{t-1}(i,j)) then
        numSameNode = numSameNode+1;
      end if
    end for
  end for
  return the ratio r = numSameNode/n;
end for
```

vertices in the same communities but a sparse link density among nodes from different communities. The modularity is defined as follows:

$$Q=\sum_{s=1}^{k}\left[\frac{l_s}{m}-\left(\frac{d_s}{2m}\right)^2\right] \tag{2}$$

where $l_s$ is the number of edges joining vertices inside the community $C_s^t$, $d_s$ is the sum of the degrees of the nodes in $C_s^t$, and $m$ is the number of edges in the network $G^t$.

In addition, NMI is employed to detect communities in [10], which can be seen by using the community label from the last snapshot. Two sets of communities $A=\{A_1, A_2,\ldots, A_{C_A}\}$ and $B=\{B_1, B_2,\ldots, B_{C_B}\}$ are given, and $C$ is the confusion matrix whose element $C_i^j$ is the number of nodes in the community $A_i \subseteq A$ that also appears in the community $B_j \subseteq B$. The definition of *NMI* is shown as follows:

$$NMI(A,B)=\frac{-2\sum_{i=1}^{c_A}\sum_{j=1}^{c_B}C_{ij}\log(C_{ij}N/C_{i.}C_{.j})}{\sum_{i=1}^{C_A}C_{i.}\log(C_{i.}/N)+\sum_{j=1}^{C_B}C_{.j}\log(C_{.j}/N)} \tag{3}$$

where $c_A(c_B)$ is the number of communities in $A(B)$, and $C_{i.}$ denotes the sum of the figures of $C$ in row $i$, and column $j$ is marked $C_{.j}$. The maximum value of *NMI* is 1, and the minimum value of NMI is 0. If $A=B$, NMI($A$, $B$)=1. On the contrary, if $A$ is different from $B$, NMI($A$, $B$)=0. The higher the NMI($CR^t$, $CR^{t-1}$), the lower the temporal cost is. NMI and $Q$ are conflicting objectives (see Supplementary material, Section III).

To use the higher-order knowledge (the community label not only from the last snapshot but also the previous snapshots), we design the new objective HoNMI. The definition of HoNMI is shown as follows:

$$HoNMI=\sum w(i)\cdot NMI(CR^i,CR^t), i=1,2,\ldots,t-1 \tag{4}$$

where $w(i)$ refers to the weight at step $i$ ($i$=1, …, $t$-1) and $\sum w(i)=1$. It can employ higher-order knowledge effectively to avoid negative transfer when the network abruptly changes. Since we found that the previous snapshots still have a high similarity with the current snapshot, HoNMI 1) adopts the majority voting principle to remove the negative transfer effect of the last snapshot and 2) makes full use of the community label information of other highly similar snapshots.

### *C. Higher-order Knowledge Transfer*

Indirect transfer learning is adopted in our algorithm. *NMI*($CR^t$, $CR^{t-1}$) measures clustering drift from time step $t$ to time step $t$-1. The *NMI* is frequently set as an objective in many studies to minimize the difference between different community structures. The hypothesis exists in those efforts: the communities should not sharply change between two successive time steps, and the evolution should be presented as a smooth transition. However, the existence of the hypothesis is a problem on its own due to the condition where clusters may change violently is not taken into account. Thus, we design a strategy introducing *HoNMI* as an objective function to use higher-order knowledge. We express first-order knowledge and higher-order knowledge with an objective, *HoNMI*. The definition of first-order knowledge and higher-order knowledge is shown as follows:

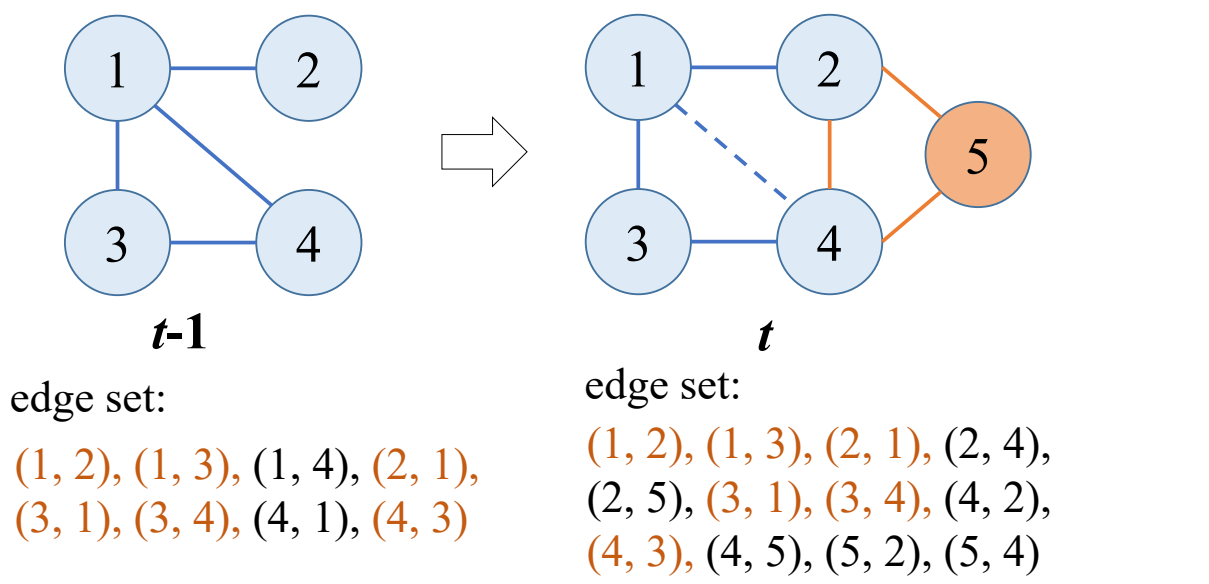

Figure 3 An example is to compute the overlapping ratio. The left is the network at time step $t$-1, and the right is the network at time step $t$. There are edge sets below the network topology, where the marked red elements are the edges that they share.

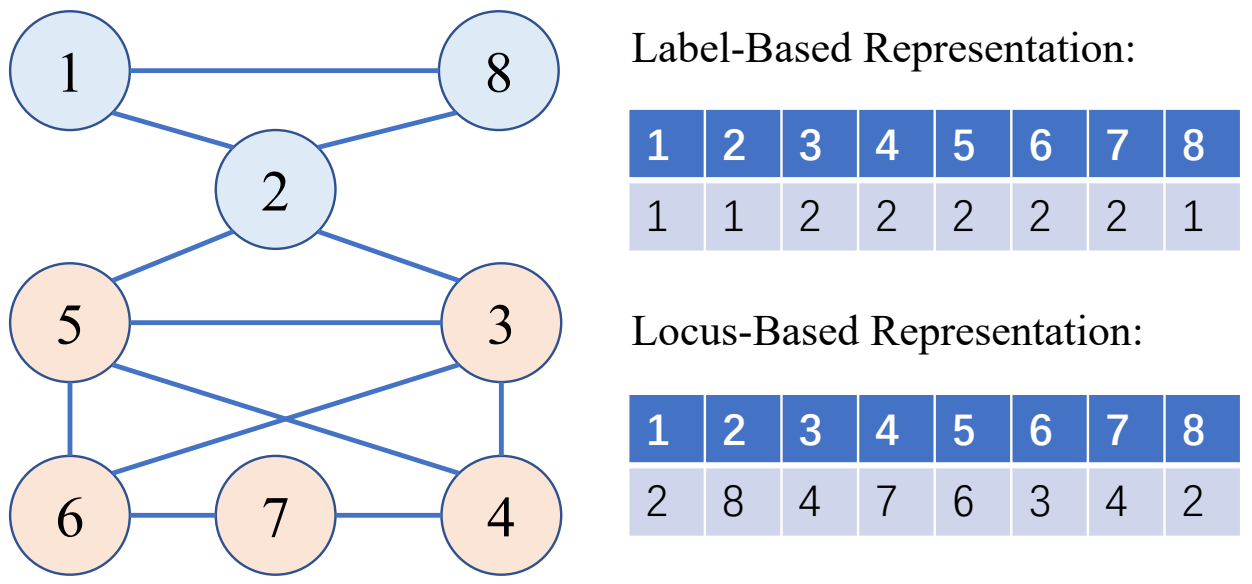

Figure 4 In encoding methods, the picture in the top right corner is the label-based representation, and the picture in the bottom right corner is the locus-based representation.

**First-order knowledge**. It computes the community label at time step $t$ using the community label obtained at time step $t$-1. The objective function *HoNMI* is equal to *NMI*($CR^{t-1}$, $CR^t$).

**Higher-order knowledge**. It computes the community labels at time step $t$ using community labels at previous time steps. The objective function *HoNMI* is equal to $\sum NMI(CR^i, CR^t)\cdot w(i)$, $i$=1, ..., $t$-1.

The overlapping ratio of the adjacency matrix at $t$ to $t$-1 is calculated using the adjacency matrices at two successive time steps. We calculate the overlapping rate as follows: ratio=numSameNode/$n$, where numSameNode is the number of the nodes that belong to two networks at the successive time steps, and $n$ is the number of nodes at time step $t$. Fig. 3 shows an example of computing the overlapping ratio. As Fig. 3 shown, the edge set at time step $t$-1 is {(1, 2), (1, 3), (1, 4), (2, 1), (3, 1), (3, 4), (4, 1), (4, 3)} and the edge set at time step $t$ is {(1, 2), (1, 3), (2, 1), (2, 4), (2, 5), (3, 1), (3, 4), (4, 2), (4, 3), (4, 5), (5, 2), (5, 4)}. There are 8 and 12 edges at time steps $t$-1 and $t$, respectively. The common edges that appear simultaneously in both networks are {(1, 2), (1, 3), (2, 1), (3, 1), (3, 4), (4, 3)}. Thus, numSameNode=6 and $n$=12 and the overlapping ratio $r$ is obtained as 0.5. Algorithm 1 describes the steps to get it.

Suppose the ratio $r \geq \sigma$, the first-order knowledge from the previous time snapshot is transferred. The ratio $r < \sigma$ demonstrates that the negative transfer may happen if we use the first-order knowledge. Thus, we select the higher-order knowledge to improve performance, where the decision-makers define the value of $\sigma$. In this way, we indirectly use the higher-order knowledge to enhance the performance of community detection at time $t$, and the failure of direct use of historical knowledge also guarantees the diversity of initializing population. In effect, the current algorithm takes advantage of only the result of the previous snapshot. However, the current network shares similarities with all the networks preceding the current time step. Therefore, all of the results of the previous networks are available.

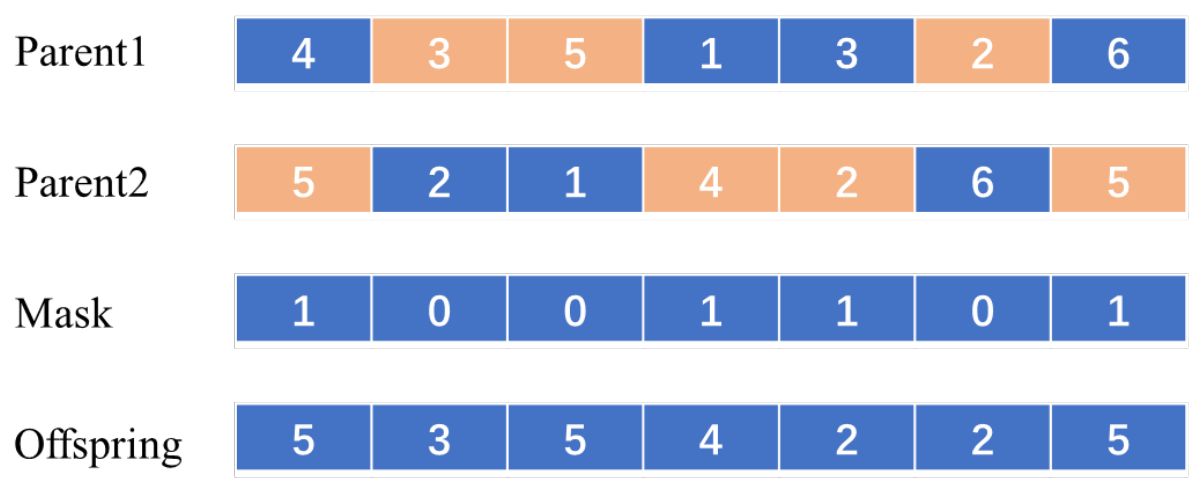

Figure 5 Crossover operator. When the mask(1)=1, the parent2(1) is selected to form the offspring(1), and then the offspring(1)=parent2(1)=5. When the mask(2)=0, the parent1(2) is selected to form the offspring(2), and then the offspring(2)=parent1(2)=3.

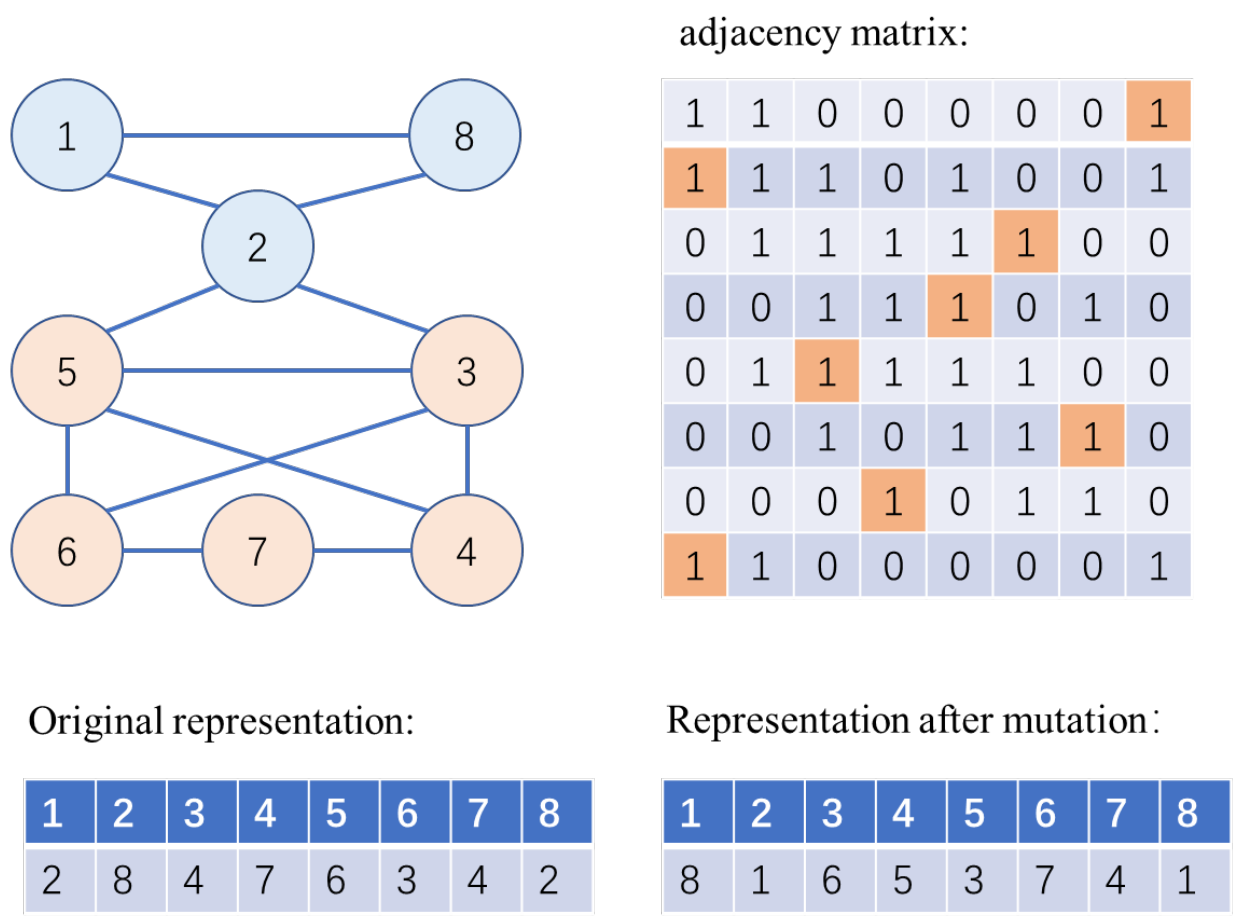

Figure 6 Mutation operator.

When the ratio $r$ is greater than the threshold $\sigma$, HoNMI=NMI($CR^t$, $CR^{t-1}$) is set as the second objective to detect communities. Furthermore, when the ratio $r$ is less than the threshold $\sigma$, we compute the *NMI* between $t$ and all time steps before time step $t$ and then add them up based on the weight $w$ to act as HoNMI, where $w$ is decided by the similarity matrix. The selection of $w$ is shown in Section IV in detail.

## D. Encoding

There are two categories representing clusters: label-based representation and locus-based representation. In label-based representation, a population consists of $N$ solutions $X=\{X_1, ..., X_N\}$, each including $n$ genes ($g_1$, $g_2$, ..., $g_n$), where $n$ is the number of nodes. If $k$ is the number of the communities, the value of $g_i$ is in the range {1, ..., $k$} that identifies the community to which node $i$ belongs. Each gene can assume allele values $g_i$ in the range {1, ..., $n$} in locus-based representation. The value of $g_i=j$ is interpreted as a connection between node $i$ and $j$. As shown in Fig. 4, nodes 1, 2, and 8 belong to the same community, and nodes 3, 4, 5, 6, and 7 belong to another community. In the label-based representation, the label of each node is the serial number of the community to which the node belongs. Thus, the label of nodes 1, 2, and 7 is 1, and the label of nodes 3, 4, 5, 6, and 7 is 2. In the locus-based representation,

**Algorithm 2: HoKT**

**Input**:
$G=\{G^1, ..., G^T\}$: the sequence of graphs;
$T$: time steps;
$w_t$: the weight for higher-order knowledge transfer at time steps $t$;
$\sigma$: the threshold;
**Output**:
**Result**: A dynamic community structure $C=\{C_1, C_2, …, C_T\}$.

```
Initialize the clustering CR^1={C_1^1, C_2^1, …, C_k^1} of the network N1 by optimizing Eq. (2).
for t = 2 to T
   Compute the overlapping ratio r of t to t-1;
      if (r ≥ σ) then
         HoNMI = NMI(t, t-1);
      else
         HoNMI = NMI(t, 1:t-1)·w_t;
      end if
   Create a random population of individuals with the number n=|V^t|;
   Decode each individual to generate the partitioning CR^t = {C_1^t, ..., C_k^t};
   Evaluate the two fitness values of the translated individuals (Q, NMI);
   while (termination criteria are not satisfied) do
      Create a population of offspring by applying the crossover and mutation operators;
      Decode each individual to generate the partitioning CR^t={C_1^t, ..., C_k^t};
      Evaluate the two fitness values of the translated individuals (Q, NMI);
      Assign a rank to each individual and sort them according to non-domination rank;
      Combine the parents and offspring into a new pool and partition it into fronts;
      Select points on the lower front and apply the mutation and crossover operator on them to create the next population;
   end while
   Return the solution CR^t={C_1^t, ..., C_k^t} with the maximum modularity value;
end for
```

the label of a node is a node number belonging to the same community as the node. As a result, the label of node 1 can be set to 8 or 2, and node 2 is selected randomly as the final label of node 1.

A sharp advantage of the locus-based representation is that the number $k$ of clusters can be automatically determined by the number of components contained in an individual and obtained by the decoding step [52]. Thus, locus-based adjacency encoding is employed in our algorithm.

## E. Genetic Operators

**Initialization.** The label of node $i$ is initialized by randomly choosing a neighbor from a data set. Each couple ($i$, $g_i$) represents a link belonging to one of the components of $G$.

**Crossover.** Two individuals are given to be used as parents, and a random binary mask is created. The gene belonging to parent one is selected when the mask is 0, and the gene belonging to parent two is selected when the mask is 1. Those genes are chosen to form the genes of the child. The crossover process is shown in Fig. 5, and mask(1)=1. As a result, the gene of parent two is selected as the gene of offspring, offspring(1)=parent2(1)=5. The second element of the mask is 0; thus, the gene of parent one is selected as the gene of offspring, offspring (2)=parent1(2)=3. The red elements are genes determined by the mask.

**Mutation.** Changing the value of the $g_i$ with one of the neighbors of node *i* randomly. An example is shown in Fig. 6. The original set of labels with locus-based representation is {2, 8, 4, 7, 6, 3, 4, 2}. The neighbors of node 1 is {2, 8} (node 1 is excluded). Then a node from the neighbor set is selected randomly as the label of node 1. Thus, node 8 is selected to form the label of node one after mutation. After performing the mutation operator, nodes 1, 6, 5, 3, 7, 4, and 1 are selected to become the labels 2, 3, 4, 5, 6, 7, and 8, respectively.

**Selection.** The binary tournament selection operator is applied in HoKT. According to the non-dominated sorting strategy in NSGA-II, we first assign a rank to each individual, sort them according to non-domination rank, and then combine the parents and offspring into a new pool and partition them into fronts. Finally, we select points on the lower front and apply the mutation and crossover operator to produce the next population.

### *F. Overall Procedure*

HoKT automatically provides a solution to determine whether to use high-order temporal smoothness or first-order temporal smoothness by obtaining the overlapping ratio of a snapshot to the previous shot. HoKT employs the nondominated sorting genetic algorithm (NSGA-II) [19] to optimize the designed objectives. It finds the communities of the first snapshot by optimizing only the first objective and generates the final clustering of the other snapshots by evaluating two objectives. We introduce two mechanisms to transfer knowledge, which are decided by the overlapping ratio of the adjacency matrix at *t* to *t*-1.

The flowchart of HoKT is shown in Fig. 7. Given a dynamic network $G=\{G^1, \ldots, G^T\}$, HoKT finds a partitioning of $G^1$ by running the genetic algorithm that optimizes only (2). We use the locus-based adjacency representation to initialize and detect the network. Since no knowledge can be used at $t = 1$, HoKT uses the genetic algorithm to detect communities with the first objective ((2)). When $t = 2$, there is only first-order knowledge. The population that is the most similar to the previous community structure is selected from the initialized population. Thus, this method extracts the feature of the previous community structure from the current network. NMI is the metric to measure how similar the community structure $CR^t$ is to the previous clustering $CR^{t-1}$. We use NMI as the objective function to select the most suitable population. The higher-order knowledge utilization is carried out when $t > 2$. The overlapping ratio r between the current network $G^t$ and the previous network $G^{t-1}$ is obtained. The fact that *r* is low means that using first-order knowledge causes negative transfer learning, so we get the combination of NMI($CR^t$, $CR^{t-1}$) and NMI($CR^t$, $CR^{t-i}$) as the objective function, where $i$=1, …, $t$-1. By this method, we can transfer higher-order features to the current detection. Moreover, when *r* is high enough, it is believed that the first-order knowledge has much helpful information. Thus, first-order knowledge is still used to transfer knowledge.

A random population with $n=|V^t|$ individuals is created for a given number of time steps. Then, for a fixed number of generations, it decodes the individuals to generate the partitioning at time step *t* and evaluates the objective values.

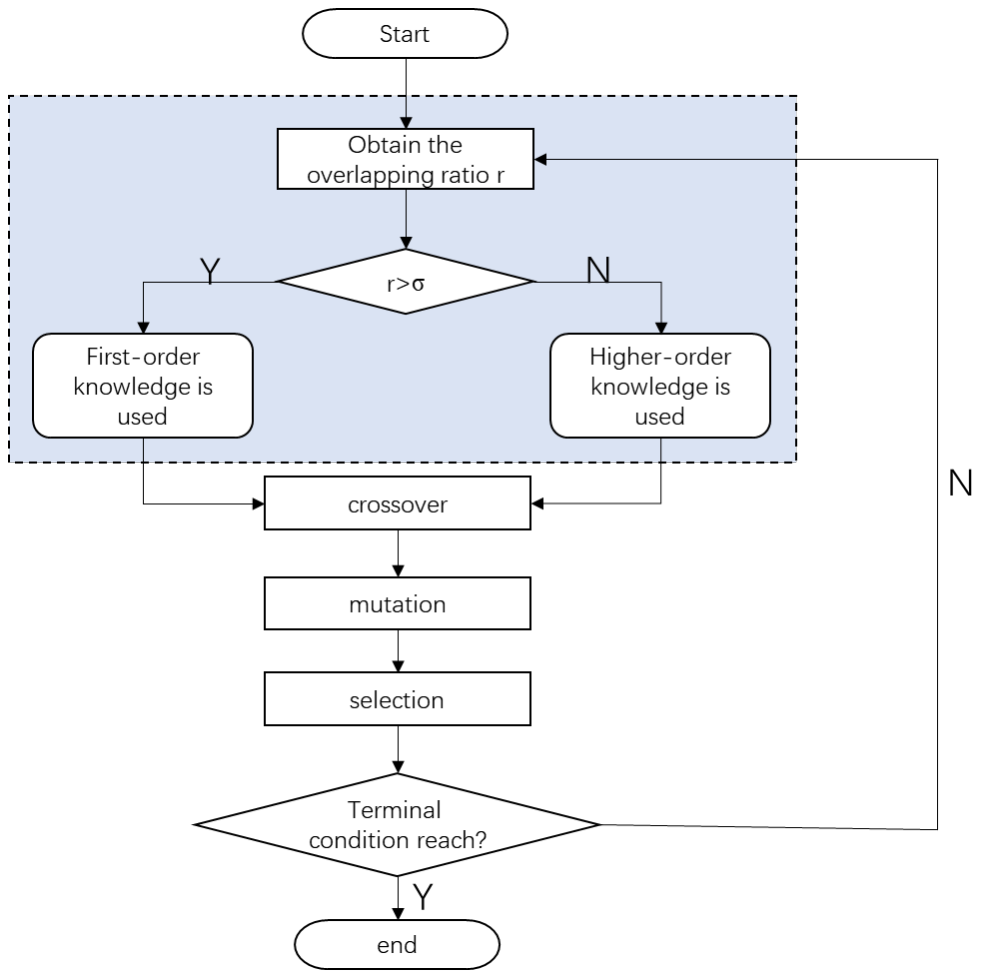

Figure 7 The flowchart of HoKT.

The crossover and mutation operators are used to create a new population. Parents and offspring are combined, and the new pool is partitioned into fronts. The selection operator is employed to assign a rank to each individual according to Pareto dominance and sorts them. The individuals with the lower rank are selected, and the crossover and mutation operators are applied to create the new population. At the end of each time step, HoKT returns a set of solutions, each corresponding to a different trade-off between (2) and (3). According to DYNMOGA [10], we choose the partitioning with the highest modularity value. Algorithm 2 describes the detailed process of HoKT.

### *G. Computational Complexity*

According to [10], the overall complexity of DYNMOGA is $O((gp\cdot logp)\times(n\cdot logn+|E|))$, where $|E|$ is the number of edges, $n$ is the number of nodes, $g$ is the number of generations, and $p$ is the population size. The biggest difference between HoKT and DYNMOGA is the selection of higher-order knowledge transfer and first-order knowledge transfer. Algorithm 1 needs $O(n^2T)$ time, which is the number of time steps. Thus, the overall complexity of HoKT is $O((gp\cdot logp)\times(n\cdot logn+|E|))$.

## IV. Experiment Results

Our experimental results show that some reported algorithms for detecting communities perform poorly when the overlapping ratio of a snapshot to the previous snapshot is not great. Moreover, to ensure the fairness of comparison, we study the effectiveness of HoKT by comparing it with the other two algorithms, including DYNMOGA and KT-MOEA/D, in terms of *NMI* and $F_1$-score. The algorithm DYNMOGA is referred to in [10], and HoKT is proposed based on DYNMOGA. Thus, DYNMOGA is run to compare with HoKT for fairness. KT-MOEA/D [26] is selected to show the improvement of HoKT.

TABLE I
Parameters of HoKT.

| Parameters | Value |
|---|---|
| Population size | 200 |
| Generations | 100 |
| Crossover Probability | 0.8 |
| Mutation Probability | 0.2 |
| The order of knowledge | 1, 2, 3 |

TABLE II
NMI AND $F_1$-SCORE FOR CELL PHONE CALL DATASET WITH MINOR CHANGES.

| Time | NMI | | | $F_1$-score | | | $w$ |
|---|---|---|---|---|---|---|---|
| | HoKT | DYNMOGA | KT-MOEA/D | HoKT | DYNMOGA | KT-MOEA/D | |
| $t$ = 1 | 0.442±0.06 | 0.349±0.12 | **0.486±0.06** | 0.414±0.09 | 0.378±0.08 | **0.431±0.08** | {} |
| $t$ = 2 | 0.744±0.00 | 0.734±0.04 | **0.756±0.05** | **0.738±0.00** | 0.706±0.06 | 0.705±0.07 | {1} |
| $t$ = 3 | **0.758±0.00** | 0.700±0.03 | 0.754±0.04 | **0.745±0.00** | 0.712±0.04 | 0.701±0.08 | {0.8, 0.2} |
| $t$ = 4 | **0.545±0.00** | 0.539±0.05 | 0.520±0.05 | **0.545±0.00** | 0.538±0.07 | 0.475±0.11 | {0.6, 0.4} |
| $t$ = 5 | **0.654±0.08** | 0.636±0.07 | 0.635±0.07 | 0.631±0.00 | **0.656±0.09** | 0.578±0.05 | {0.8, 0.2} |
| $t$ = 6 | **0.739±0.00** | 0.710±0.05 | 0.736±0.05 | **0.733±0.07** | 0.692±0.06 | 0.726±0.04 | {0.7, 0.2, 0.1} |
| $t$ = 7 | **0.654±0.00** | 0.621±0.06 | 0.622±0.09 | **0.651±0.00** | 0.587±0.11 | 0.570±0.09 | {0.9, 0.1} |
| $t$ = 8 | **0.713±0.06** | **0.717±0.05** | 0.645±0.08 | **0.710±0.07** | 0.696±0.05 | 0.636±0.10 | {0.8, 0.2} |
| $t$ = 9 | **0.618±0.00** | 0.589±0.07 | 0.602±0.06 | **0.614±0.00** | 0.517±0.09 | 0.598±0.07 | {0.7, 0.2, 0.1} |
| $t$ = 10 | 0.625±0.00 | 0.606±0.03 | **0.638±0.04** | **0.607±0.00** | 0.588±0.06 | 0.598±0.04 | {0.7, 0.3} |
| win/tie/loss | - | 1/0/9 | 3/0/7 | - | 1/0/9 | 1/0/9 | |

TABLE III
NMI FOR CELL WITH GREAT CHANGES.

| Time | HoKT | DYNMOGA | $w$ |
|---|---|---|---|
| $t$ = 1 | 0.3594±0.01 | **0.4019±0.00** | {} |
| $t$ = 3 | **0.6653±0.00** | 0.6245±0.01 | {1} |
| $t$ = 5 | 0.6121±0.00 | **0.6246±0.00** | {1} |
| $t$ = 7 | **0.6215±0.00** | 0.5659±0.00 | {0.7, 0.2, 0.1} |
| $t$ = 9 | **0.5687±0.00** | 0.5361±0.00 | {0.7, 0.2, 0.1} |
| win/tie/loss | - | 2/0/3 | |

### A. Experimental Setup

*1)* ***Performance Metric***. This paper employs two metrics to evaluate the performance of community detection: NMI and $F_1$-score. The NMI in the experimental results refers to the NMI($CR^t$, $CR^t_{real}$), where $CR^t_{real}$ is the real community label at time step $t$. $F_1$-score, a metric of classification problem, is the harmonic mean of precision and recall. $F_1$-score is defined as follows,

$$F_1\text{-score} = 2\times \frac{TP}{(TP+FN)(TP+FP)} \quad (3)$$

where $TP$ is the number of labels predicted correctly, $FN$ is the number of labels predicted to be negative but positive in the real case, and $FP$ is the number of labels predicted positive but negative in the real case.

*2)* ***Baseline Methods***: When considering the great network change scenario, the main goal of this paper is to verify that the proposed higher-order knowledge transfer strategy can effectively balance the effects of positive and negative transfer. Most of the dynamic multi-objective community detection methods introduced in Section III.B are variants of DYNMOGA, and they do not consider how to eliminate the impact of great network changes. Therefore, we only choose DYNMOGA as the benchmark. At the same time, there are also non-evolutionary algorithms, as mentioned in [18], [41], [45]. They do not consider the impact of great network changes, and they are also different from the optimizers of our scheme. Comparing with them cannot verify the motivation of our scheme, and it is a meaningless and unfair comparison. We can transfer the ideas of this paper to other dynamic community detection methods or dynamic optimization problems; however, doing so deviates from our original intention, which can be regarded as future work.

Most dynamic multi-objective algorithms based on transfer learning introduced in Section III.C are solutions for continuous optimization problems, especially their knowledge transfer strategies that are difficult to solve the problem of dynamic community detection directly. Meanwhile, these schemes rarely consider the impact of great changes. We find that a knee point-based transfer learning method called KT-MOEA/D [26] can handle the problem of dynamic community detection. Therefore, we selected KT-MOEA/D as a representative algorithm to verify the effectiveness of our transfer scheme.

DYNMOGA and KT-MOEA/D [26] are employed as baselines. KT-MOEA/D and DYNMOGA optimize (2) and (3), using only first-order temporal smoothness. Moreover, KT-MOEA/D employs transfer learning to initialize the population of the subsequent snapshots.

*3)* ***Parameters***. Setting parameters is a significant challenge for whatever algorithms we use. HoKT has five parameters to control: population size, generations, crossover probability, mutation probability, the order of knowledge, $w$, and $\sigma$. The parameters are shown in Table I. The crossover, mutation, and selection operators of KT-MOEA/D and DYNMOGA are the same as HoKT. For each dataset, 30 independent runs are performed for all methods, and the Wilcoxon rank-sum test is employed to test the significance of the results.

### B. Cell Phone Call

The first dataset we applied is a real-life dynamic network, the Cell Phone Call dataset. This dataset consists of cell phone call records for ten days in June 2006, in which each of the nodes represents cell phones, and an edge increases with the connection between the two cell phones. There are 400 nodes and five core points in this network. The comparison between DYNMOGA and our algorithm shows the advantage of higher-order temporal smoothness, as shown in Table II. The best NMI or $F_1$-score values are marked with bold black font. In addition, we summarize the significance test results with the NMI and $F_1$-score values at the bottom row. The similarity for the Cell Phone Call dataset is shown in Fig. 8. The element $(i, j)$ is the overlapping ratio between the $i$th and $j$th snapshots.

HoKT tends to have the best performance over snapshots 3-9. This phenomenon is because the longer the time, the higher-order knowledge is available. HoKT mainly employs second-order knowledge and third-order knowledge. We set $w$={0.6, 0.4}, {0.7, 0.3}, {0.8, 0.2}, or {0.9, 0.1} to transfer the second-order knowledge and $w$={0.5, 0.3, 0.2}, {0.6, 0.3, 0.1}, or {0.7, 0.2, 0.1} to transfer the third-order knowledge. We select the best results for the final presentation.

No other knowledge is used at the first step, and all three algorithms detect communities with static methods. As a result, HoKT gets a poor grade. At time step 2, HoKT, DYNMOGA, and KT-MOEA/D use first-order knowledge. The NMI of HoKT is worse than KT-MOEA/D but better than DYNMOGA. At time step 3, the second-order knowledge is used, and $w$ = {0.8, 0.2}, HoKT shows its advantages and gets the best NMI and $F_1$-score. The overlapping ratio $r_{13}$ of snapshots between $t$=1 and $t$=3 is 0.7506 and $r_{23}$=0.7600. The difference between those two $r$ values is noticeable, and $r_{23}$ is greater than $r_{13}$. Thus, the weight for $t$=2 is set to 0.8, and the weight for $t$=1 is 0.2.

TABLE IV
NMI AND $F_1$-SCORE FOR ENRON DATASET WITH MINOR CHANGES.

| Time | NMI | | | $F_1$-score | | | $w$ |
|---|---|---|---|---|---|---|---|
| | HoKT | DYNMOGA | KT-MOEA/D | HoKT | DYNMOGA | KT-MOEA/D | |
| $t$ = 1 | 0.4185±0.08 | 0.4058±0.08 | **0.4464±0.13** | **0.4163±0.10** | 0.3903±0.06 | 0.4125±0.15 | {} |
| $t$ = 2 | **0.7437±0.03** | 0.6928±0.07 | 0.7128±0.07 | **0.7437±0.03** | 0.6864±0.08 | 0.7101±0.07 | {1} |
| $t$ = 3 | **0.7969±0.05** | 0.7197±0.04 | 0.6999±0.04 | **0.7969±0.05** | 0.7148±0.03 | 0.6925±0.06 | {0.9, 0.1} |
| $t$ = 4 | **0.5688±0.09** | 0.5233±0.03 | 0.4944±0.04 | **0.5698±0.08** | 0.5048±0.05 | 0.4837±0.06 | {0.8, 0.2} |
| $t$ = 5 | **0.6959±0.05** | 0.6241±0.08 | 0.6116±0.06 | **0.6959±0.05** | 0.6107±0.08 | 0.5886±0.06 | {0.9, 0.1} |
| $t$ = 6 | **0.7512±0.04** | 0.7269±0.05 | 0.6827±0.05 | **0.7512±0.04** | 0.7231±0.05 | 0.6801±0.05 | {0.9, 0.1} |
| $t$ = 7 | **0.6548±0.06** | 0.6070±0.08 | 0.5754±0.06 | **0.6472±0.06** | 0.5830±0.09 | 0.5591±0.06 | {0.8, 0.2} |
| $t$ = 8 | 0.6987±0.03 | **0.7063±0.05** | 0.6812±0.03 | 0.6922±0.05 | **0.7053±0.05** | 0.6818±0.03 | {0.8, 0.2} |
| $t$ = 9 | **0.6015±0.05** | 0.5924±0.05 | 0.5991±0.04 | **0.6062±0.05** | 0.5647±0.07 | 0.5794±0.06 | {0.5, 0.3, 0.2} |
| $t$ = 10 | **0.6432±0.08** | 0.6013±0.04 | 0.5780±0.03 | **0.6432±0.07** | 0.5867±0.04 | 0.5825±0.03 | {0.6, 0.3, 0.1} |
| $t$ = 11 | **0.6549±0.08** | 0.6531±0.07 | 0.6115±0.05 | **0.6690±0.06** | 0.6358±0.08 | 0.6046±0.05 | {0.7, 0.3} |
| $t$ = 12 | **0.6055±0.04** | 0.5593±0.09 | 0.5923±0.07 | **0.6112±0.07** | 0.5284±0.09 | 0.5375±0.11 | {0.7, 0.2, 0.1} |
| win/tie/loss | - | 1/0/11 | 1/0/11 | - | 1/0/11 | 0/0/12 | |

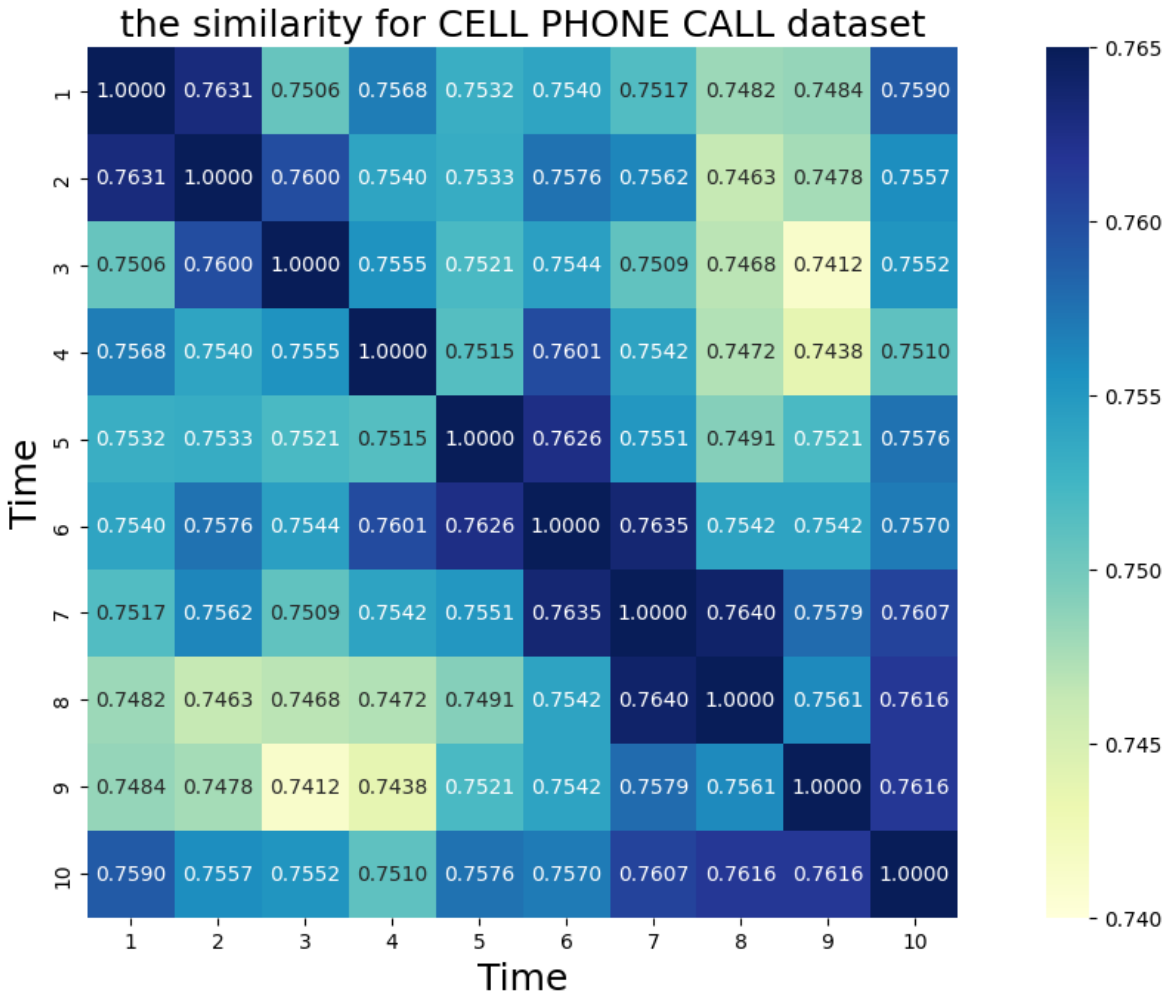

Figure 8 The similarity for Cell Phone Call dataset.

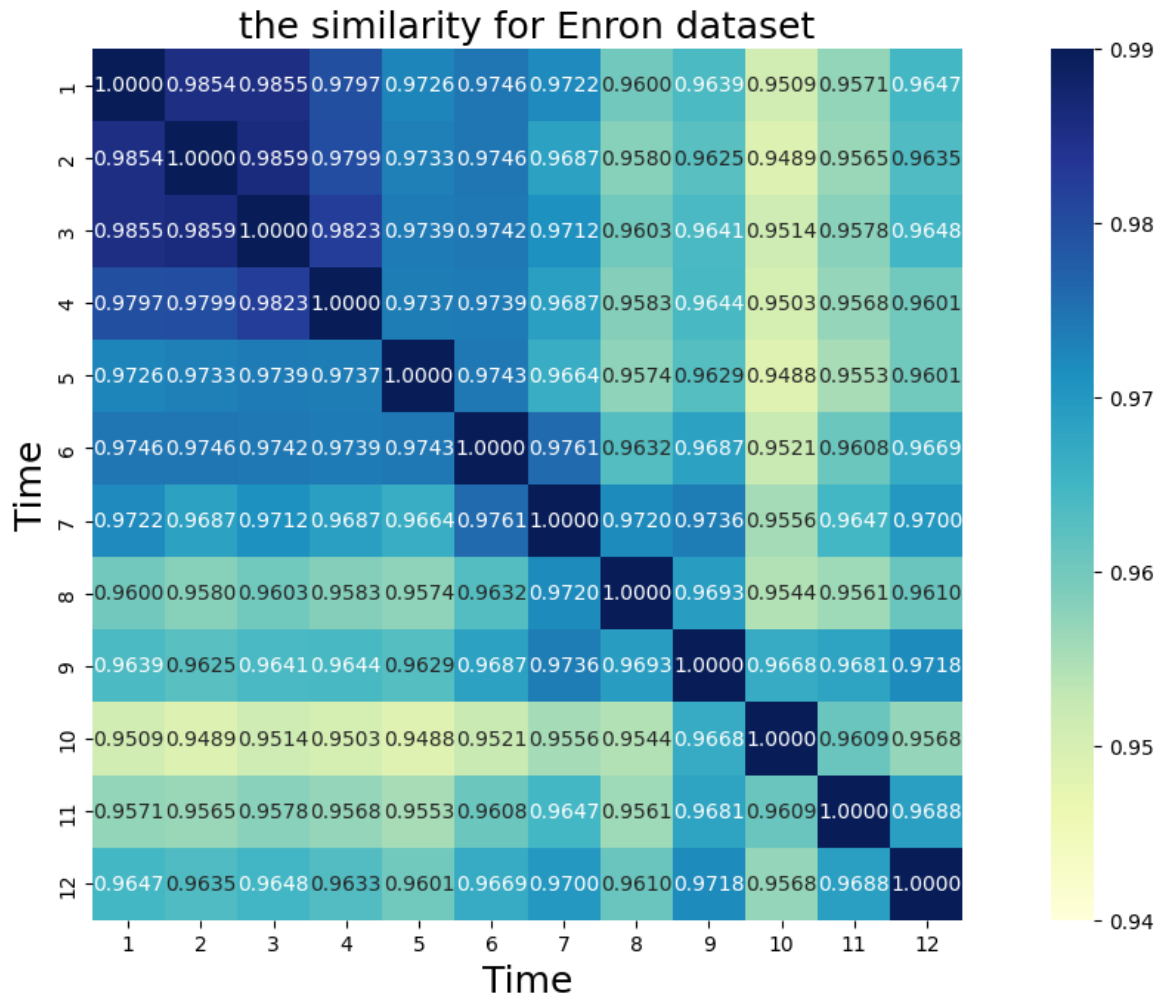

Figure 9 The similarity for Enron dataset.

TABLE V
NMI FOR ENRON WITH GREAT CHANGES.

| Time | HoKT | DYNMOGA |
|---|---|---|
| $t$ = 1 | 0.4314±0.05 | **0.4786±0.10** |
| $t$ = 3 | **0.7060±0.07** | 0.6509±0.04 |
| $t$ = 5 | **0.6469±0.05** | 0.6264±0.07 |
| $t$ = 7 | **0.6002±0.08** | 0.5846±0.06 |
| $t$ = 9 | **0.5583±0.06** | 0.4822±0.05 |
| $t$ = 11 | **0.5954±0.10** | 0.5717±0.07 |
| win/tie/loss | - | 1/0/5 |

The overlapping ratio $r_{34}$ of snapshots between $t$=4 and $t$=3 is 0.7555 and $r_{24}$=0.7540. The difference between them is small and only 0.0015. Thus, the second-order knowledge is used, and $w$={0.6, 0.4} at time step 4. At time step 5, the network at $t$=5 is more similar to the network at $t$=4 than at $t$=3, 2, and 1. The weight for the fourth snapshot is greater than the third snapshot. Thus, the second-order knowledge is only used, and $w$={0.8, 0.2}. At time step 6, the third-order knowledge is used and $w$={0.7, 0.2, 0.1}. At time step 7, $r_{67}$ is 0.7635, $r_{57}$ is 0.7551, and the overlapping ratios between the seventh snapshot and other snapshots are smaller than $r_{57}$. Thus, we select the second-order knowledge, and $w$={0.9, 0.1}. Similarly, at time step 8, the second-order knowledge is used, and $w$={0.8, 0.2} due to $r_{78}$=0.7640 and $r_{68}$=0.7542. The large overlapping ratio illustrates more helpful knowledge in this snapshot for the following snapshot. At time step 9, HoKT uses the third-order knowledge and $w$={0.7, 0.2, 0.1}. At time step 10, HoKT uses the second-order knowledge and $w$={0.7, 0.3}. At time step 10, the network does not abruptly change; as a result, the first-order temporal smoothness is used, and the NMI of the three algorithms is similar. In terms of the NMI metric, HoKT outperforms or matches DYNMOGA in all cases and loses three times to KT-MOEA/D. In terms of the $F_1$-score metric, HoKT loses once to KT-MOEA/D and DYNMOGA. HoKT prefers to obtain a high $F_1$-score during steps 2-4 and 6-10.

To obtain results in the network with great changes, we run HoKT and DYNMOGA in the first, third, fifth, seventh, and ninth-time steps of the Cell Phone Call dataset, which is displayed in Table III. HoKT achieves the best solutions compared with DYNMOGA. It is noted that HoKT selects poor results with the static method at the first step in order to demonstrate its effectiveness. However, HoKT uses third-order knowledge and $w$={0.7, 0.2, 0.1} when $t$=7 and 9. It leads to a decrease in negative transfer learning of lower-order knowledge. At time step 5, since the network structure changes slightly, DYNMOGA obtains the better NMI value. The problem needs to be improved, i.e., how to obtain good results with little time steps.

TABLE VI
NMI AND F1-SCORE FOR SYNFIX DATASET WHEN $Z$=5 WITH MINOR CHANGES.

| time steps | NMI | | | $F_1$-score | | | $w$ |
|---|---|---|---|---|---|---|---|
| | HoKT | DYNMOGA | KT-MOEA/D | HoKT | DYNMOGA | KT-MOEA/D | |
| $t$ = 1 | **0.9793±0.02** | 0.9600±0.04 | 0.9574±0.01 | **0.9771±0.03** | 0.9600±0.04 | 0.9574±0.01 | {} |
| $t$ = 2 | **1.0000±0.00** | 0.9841±0.02 | 0.9862±0.01 | **1.0000±0.00** | 0.9841±0.02 | 0.9862±0.01 | {1} |
| $t$ = 3 | **0.9975±0.01** | **0.9975±0.01** | **0.9975±0.01** | **1.0000±0.00** | 0.9975±0.01 | 0.9975±0.01 | {0.7, 0.2, 0.1} |
| $t$ = 4 | **1.0000±0.00** | 0.9901±0.02 | **1.0000±0.00** | **1.0000±0.00** | 0.9901±0.02 | **1.0000±0.00** | {0.6, 0.3, 0.1} |
| $t$ = 5 | **1.0000±0.00** | 0.9975±0.01 | 0.9975±0.01 | **1.0000±0.00** | 0.9975±0.01 | 0.9975±0.01 | {0.9, 0.1} |
| $t$ = 6 | **0.9979±0.01** | 0.9899±0.01 | 0.9975±0.01 | **0.9979±0.01** | 0.9899±0.01 | 0.9975±0.01 | {0.6, 0.4} |
| $t$ = 7 | **1.0000±0.00** | 0.9950±0.01 | 0.9970±0.01 | **1.0000±0.00** | 0.9950±0.01 | 0.9970±0.01 | {0.9, 0.1} |
| $t$ = 8 | **0.9975±0.01** | 0.9854±0.01 | 0.9929±0.01 | **0.9975±0.01** | 0.9854±0.01 | 0.9929±0.01 | {0.7, 0.2, 0.1} |
| $t$ = 9 | **0.9954±0.01** | 0.9848±0.01 | 0.9779±0.02 | **0.9954±0.01** | 0.9846±0.02 | 0.9779±0.02 | {0.5, 0.3, 0.2} |
| $t$ = 10 | **1.0000±0.00** | **1.0000±0.00** | 0.9922±0.01 | **1.0000±0.00** | **1.0000±0.00** | 0.9922±0.01 | {0.5, 0.3, 0.2} |
| win/tie/loss | - | 0/2/8 | 0/2/8 | - | 0/1/9 | 0/1/9 | |

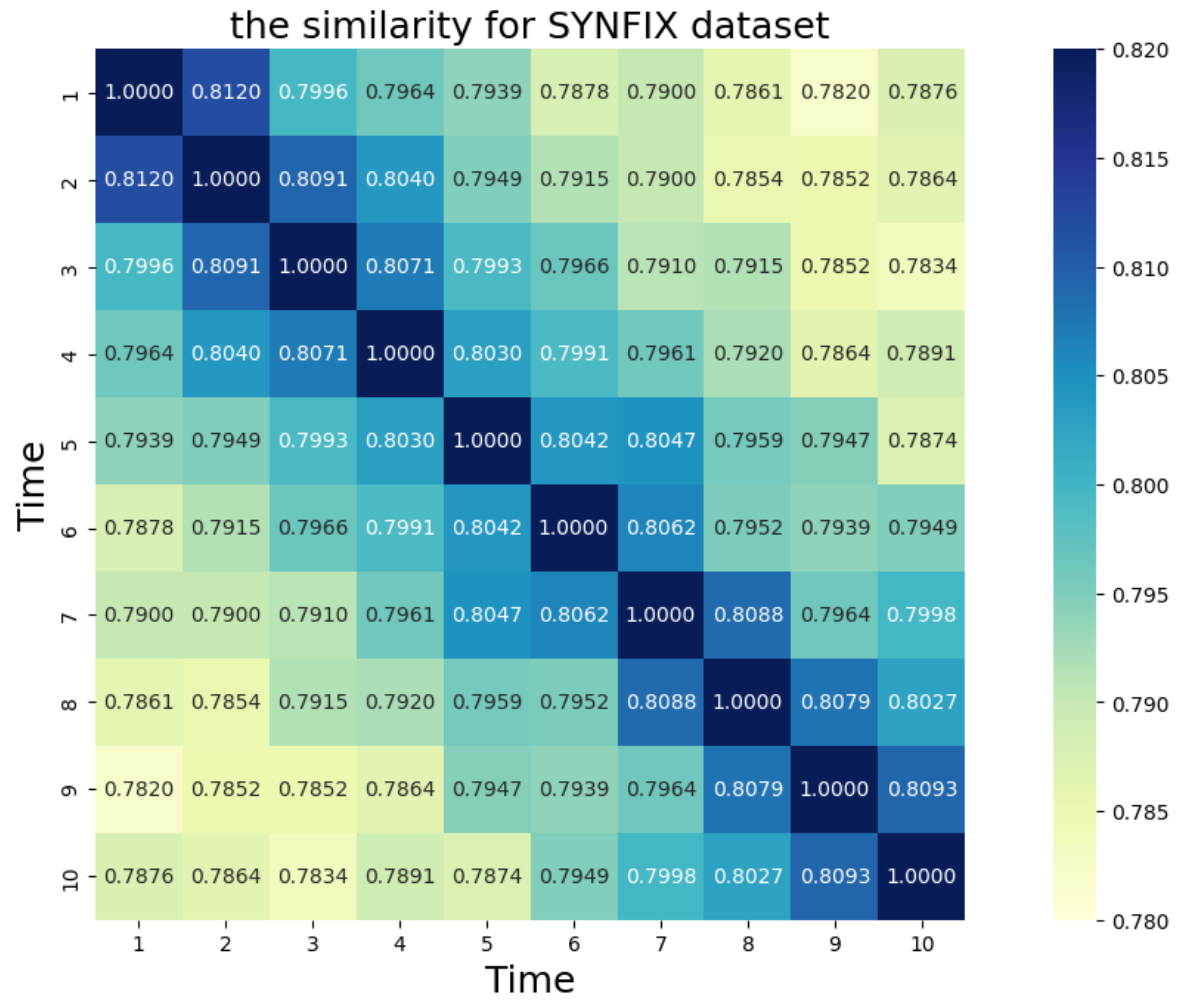

Figure 10 The similarity for SYNFIX dataset.

TABLE VII
NMI FOR SYNFIX WHEN Z=5 WITH GREAT CHANGES.

| Time | HoKT | DYNMOGA | $w$ |
|---|---|---|---|
| $t$ = 1 | **0.9707±0.04** | 0.9378±0.04 | {} |
| $t$ = 3 | **0.9950±0.02** | 0.9856±0.01 | {1} |
| $t$ = 5 | **0.9975±0.01** | 0.9925±0.02 | {0.9, 0.1} |
| $t$ = 7 | **0.9979±0.01** | 0.9859±0.02 | {0.7, 0.3} |
| $t$ = 9 | **0.9909±0.01** | 0.9861±0.02 | {0.7, 0.3} |
| win/tie/loss | - | 0/0/5 | |

TABLE VIII
NMI FOR FOUR DATASETS WITH GREAT CHANGES.

| | Time | HoKT | DYNMOGA | KT-MOEA/D | $w$ |
|---|---|---|---|---|---|
| 1 | $t$ = 1 | 0.9520±0.0185 | 0.9560±0.0173 | **0.9649±0.0107** | {} |
| | $t$ = 3 | **0.9971±0.0047** | 0.9940±0.0024 | 0.9942±0.0021 | {1} |
| | $t$ = 5 | **0.9959±0.0024** | **0.9962±0.0011** | 0.9939±0.0040 | {0.8,0.2} |
| 2 | $t$ = 1 | **0.9612±0.0143** | **0.9721±0.0281** | **0.9714±0.0163** | {} |
| | $t$ = 3 | **0.9958±0.0048** | 0.9922±0.0054 | 0.9734±0.0023 | {1} |
| | $t$ = 5 | **0.9970±0.0044** | 0.9815±0.0072 | 0.9335±0.0049 | {0.9,0.1} |
| 3 | $t$ = 1 | 0.9494±0.0142 | **0.9649±0.0102** | 0.9492±0.0195 | {} |
| | $t$ = 3 | **0.8679±0.0018** | **0.8675±0.0009** | 0.8522±0.0062 | {1} |
| | $t$ = 5 | **0.6990±0.0023** | 0.6975±0.0017 | 0.6905±0.0082 | {0.9,0.1} |
| 4 | $t$ = 1 | **0.9556±0.0227** | **0.9637±0.0048** | **0.9588±0.0082** | {} |
| | $t$ = 3 | **0.9926±0.0046** | **0.9931±0.0075** | 0.9905±0.0023 | {1} |
| | $t$ = 5 | **0.9833±0.0081** | 0.9760±0.0106 | **0.9808±0.0082** | {0.7,0.3} |
| win/tie/loss | | - | 1/6/5 | 1/3/8 | - |

TABLE IX
$F_1$-SCORE FOR FOUR DATASETS WITH GREAT CHANGES.

| | Time | HoKT | DYNMOGA | KT-MOEA/D | $w$ |
|---|---|---|---|---|---|
| 1 | $t$ = 1 | 0.9520±0.0074 | **0.9615±0.0173** | 0.9550±0.0214 | {} |
| | $t$ = 3 | **0.9974±0.0030** | 0.9920±0.0077 | 0.9921±0.0102 | {1} |
| | $t$ = 5 | **0.9959±0.0026** | **0.9964±0.0009** | 0.9933±0.0054 | {0.8,0.2} |
| 2 | $t$ = 1 | 0.9535±0.0173 | 0.9520±0.0208 | **0.9714±0.0137** | {} |
| | $t$ = 3 | **0.9972±0.0042** | **0.9965±0.0043** | 0.9928±0.0064 | {1} |
| | $t$ = 5 | **0.9960±0.0036** | 0.9811±0.0113 | 0.9927±0.0052 | {0.9,0.1} |
| 3 | $t$ = 1 | **0.9494±0.0241** | **0.9550±0.0034** | **0.9492±0.0332** | {} |
| | $t$ = 3 | 0.8512±0.0018 | 0.8507±0.0003 | **0.8778±0.0677** | {1} |
| | $t$ = 5 | 0.6560±0.0020 | 0.6547±0.0015 | **0.7226±0.1551** | {0.9,0.1} |
| 4 | $t$ = 1 | 0.9480±0.0231 | **0.9637±0.0048** | 0.9588±0.0079 | {} |
| | $t$ = 3 | **0.9922±0.0042** | **0.9931±0.0075** | 0.9903±0.0028 | {1} |
| | $t$ = 5 | **0.9807±0.0089** | 0.9734±0.0138 | 0.9789±0.0121 | {0.7,0.3} |
| win/tie/loss | | - | 2/7/3 | 4/2/6 | - |

### C. Enron

The second dataset is also a real-life network that collects email records of a company in the U.S. from 1999 to 2002, called the Enron email dataset, which provides real-world data that is arguable of the same kind as data from Echelon intercepts--a set of messages about a wide range of topics, from a large group of people who do not form a closed set [55]. The network we used consists of 50000 messages among 151 users. The similarity for the Enron dataset is shown in Fig. 9.

Table IV shows the NMI and $F_1$-score values of HoKT, KT-MOEA/D, and DYNMOGA. HoKT has a better NMI value than other algorithms in steps 2-12. As can be observed from the significant test results, HoKT obtains a competitive average NMI over DYNMOGA and KT-MOEA/D on 11 timesteps. In time step 1, we still detect communities by the static method, which is the same as KT-MOEA/D and DYNMOGA. Moreover, we select a worse result for HoKT to show its effectiveness better. We select weights according to the similarity rate between the networks.

HoKT can obtain the optimal solutions in most of the snapshots. KT-MOEA/D gets solutions with lower accuracy and more significant time cost due to its transfer strategy, which is more suitable for continuous optimization problems. Although the NMI of KT-MOEA/D is superior at the first step, it obtains worse solutions than the other two algorithms because of its transfer strategy. HoKT reduces the risk of negative transfer learning when the network changes significantly. HoKT is superior to the other two algorithms in terms of the $F_1$-score at 11-time steps. DYNMOGA obtains the better partitions only in one time step.

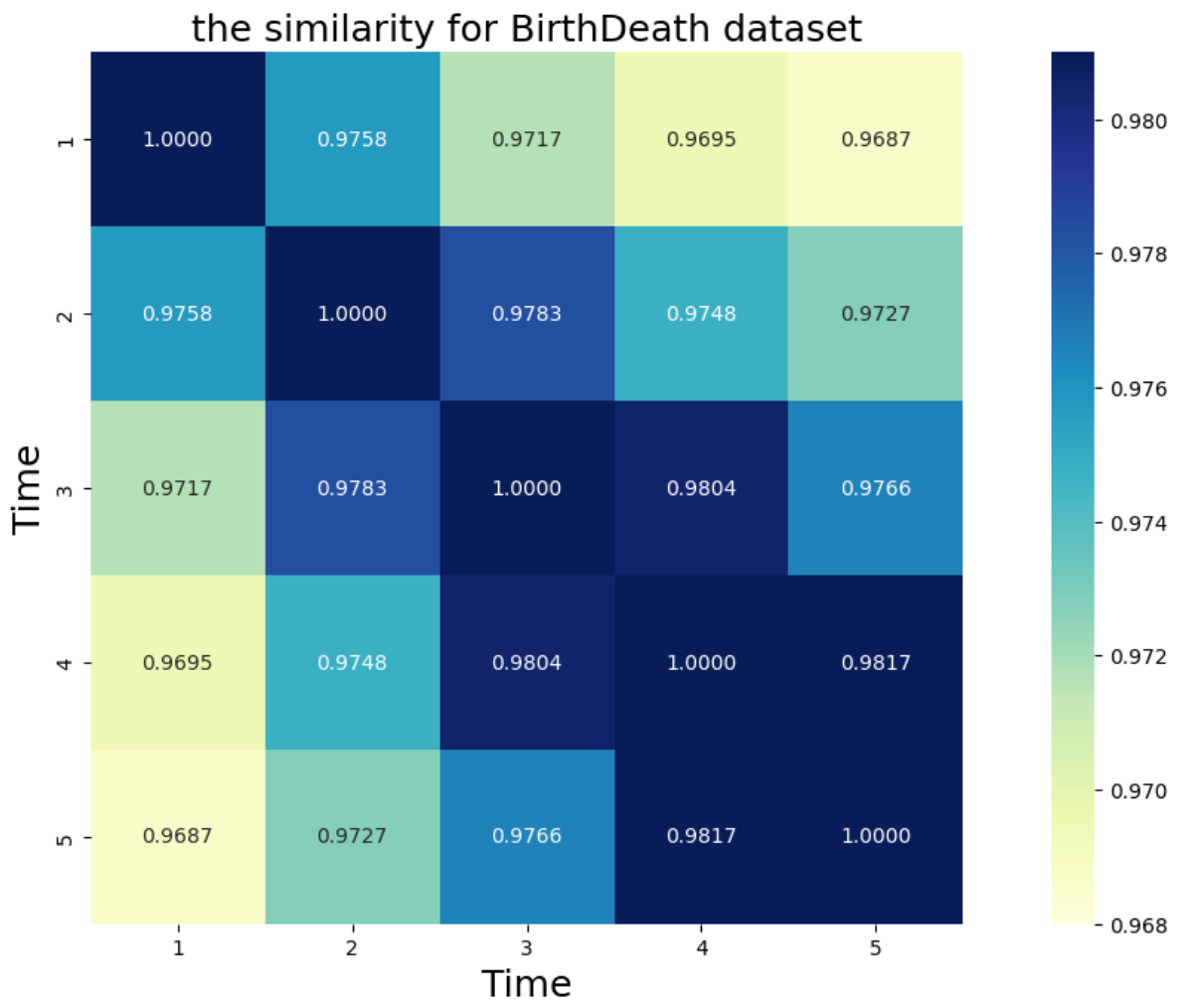

Figure 11 The similarity for Birth and Death dataset.

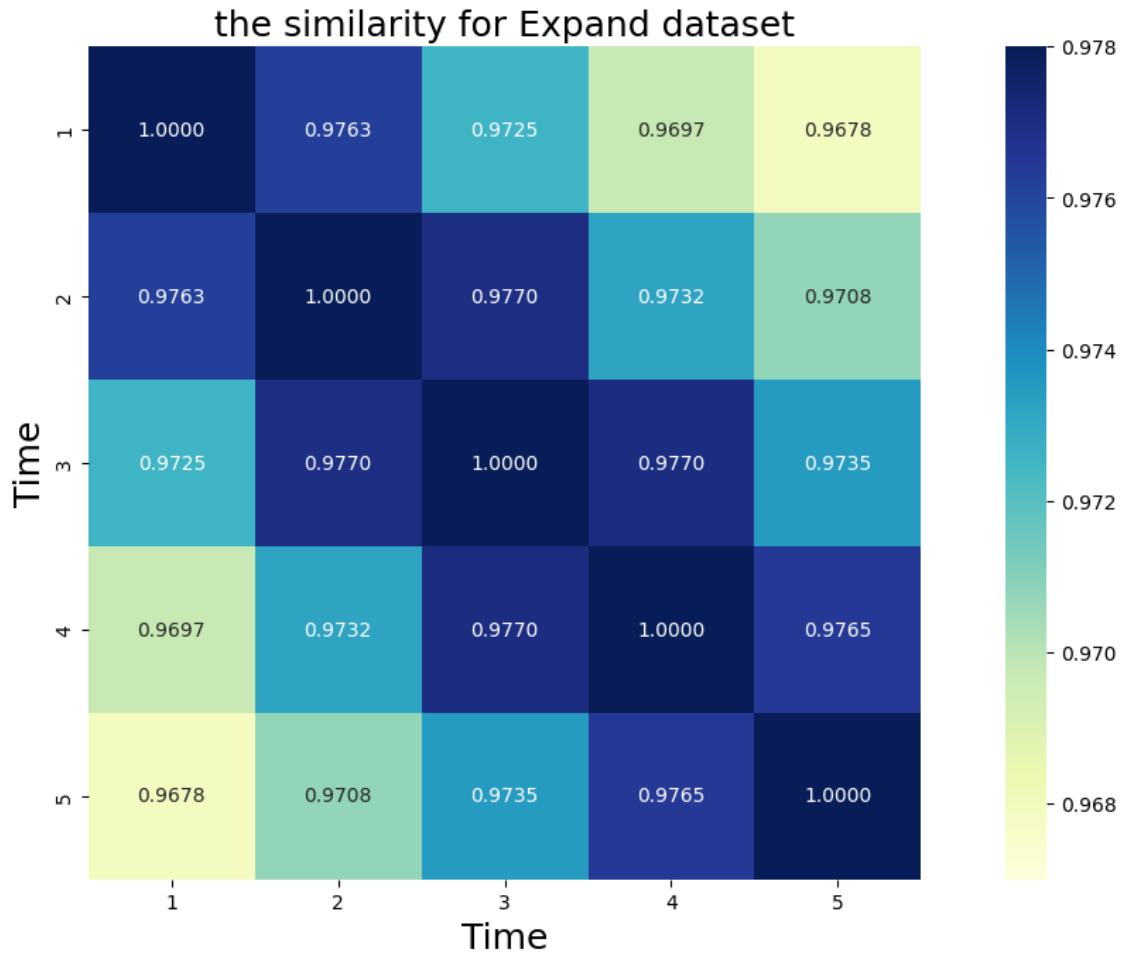

Figure 12 The similarity for Expansion and Contraction dataset.

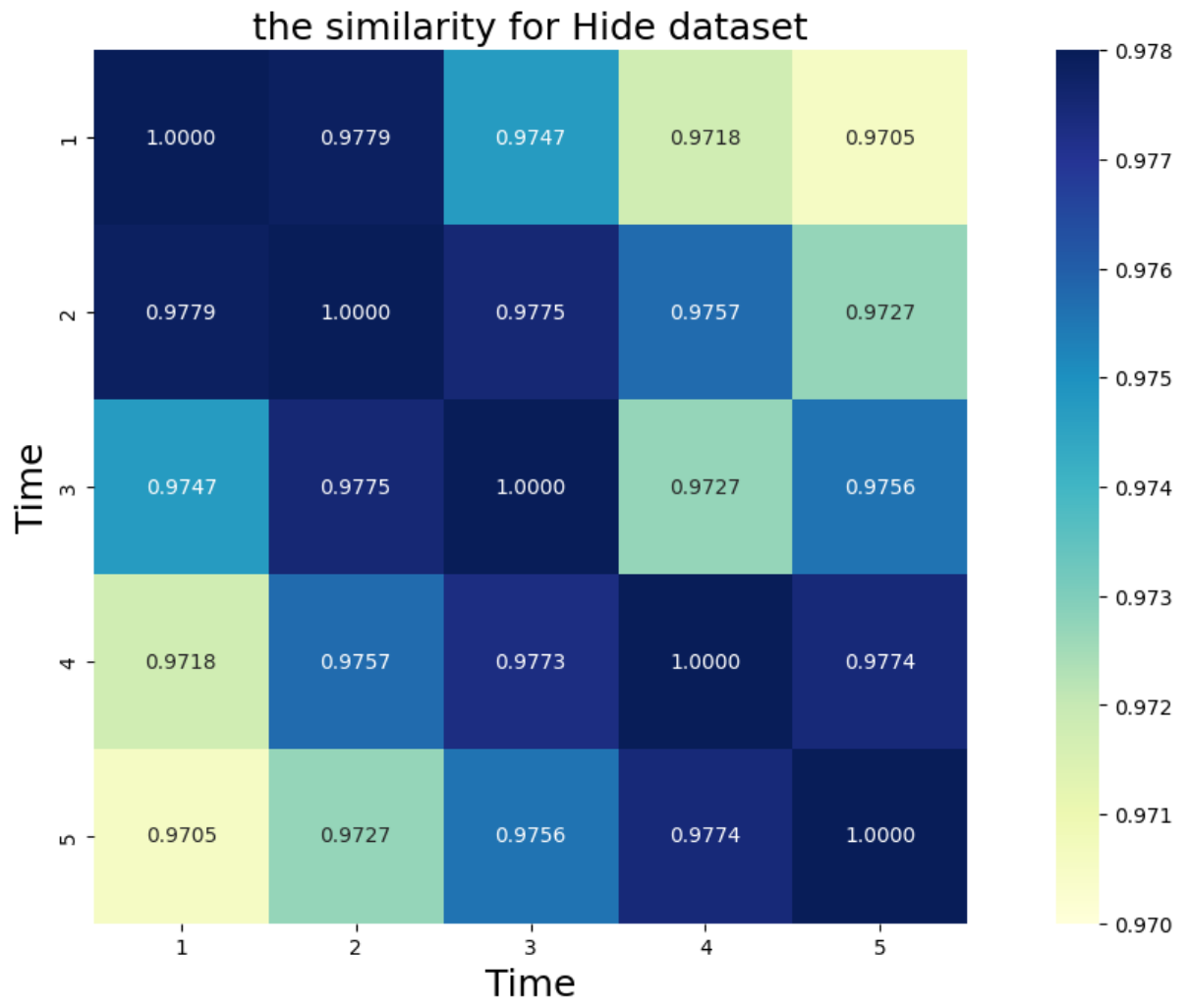

Figure 13 The similarity for Intermittent Communities dataset.

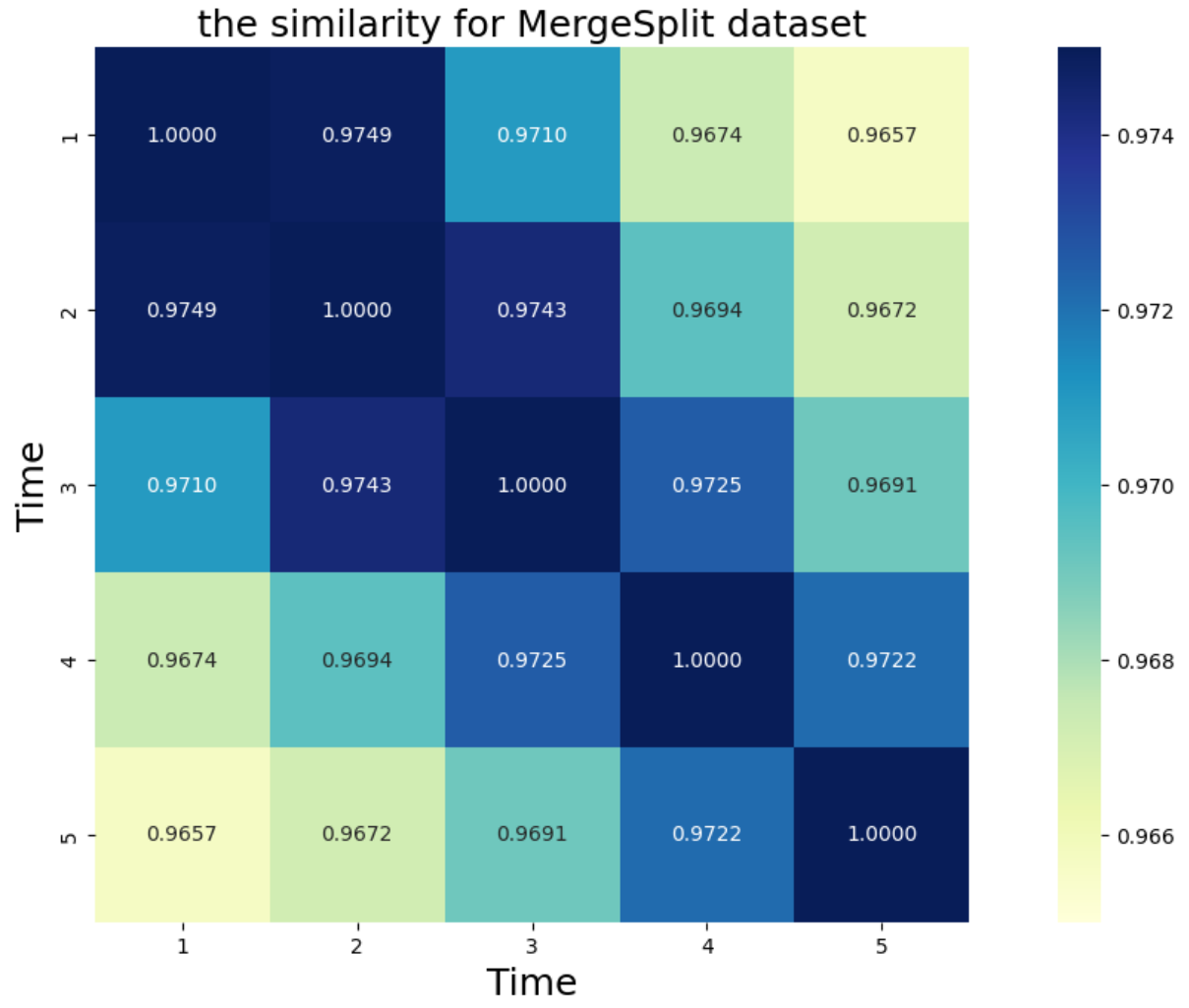

Figure 14 The similarity for Merging and Splitting dataset.

Table V shows the result in the analog change networks. HoKT is superior to DYNMOGA at time steps 3, 5, 7, 9, and 11. When $t$=1, since there is no helpful information, the detecting strategy of HoKT is equal to the process of DYNMOGA. HoKT selects a worse solution set. The value of $w$ for each time step is shown in Table V. DYNMOGA obtains the worst partitions at all steps in Enron with great changes. It indicates that DYMOGA has a poor upper limit in the network with great changes.

### *D. SYNFIX*

This section compares HoKT with KT-MOEA/D and DYNMOGA on a dataset with an unchanged number of communities called SYNFIX. The network contains four communities, each of which holds 32 nodes. Every node connects with the other 16 nodes and uses a number $z$ of links in conjunction with nodes else. Three nodes are selected and moved to the other three communities at each time step. Fig. 10 shows the similarity for the SYNFIX dataset.

Table VI shows NMI and $F_1$-score for SYNFIX when $z$=5. When $t$=2, the first-order knowledge is used. The superior performances of HoKT compared to DYNMOGA and KT-MOEA/D are on 8-time steps. The similarity rates of networks on this dataset increase over time. When $t$=9 and 10, all of the third-order knowledge is useful based on the similarities of the networks, and the weight is close.

As shown in Table VI, HoKT performs better than the other algorithms in all eight snapshots regarding NMI. NMI of HoKT is the same as that of DYNMOGA and KT-MOEA/D at time steps 3, 4, and 10. Although KT-MOEA/D has good performance at time step 4, it needs extra memory to store the knee points due to its specific transfer learning. Moreover, all the solutions obtained by HoKT when $t$=2, 3, 4, 5, 7, and 10 are equal to the ground truth. It is shown that our algorithm can achieve the best and most stable results in the SYNFIX dataset.

The NMI values in SYNFIX dataset with great changes is shown in Table VII. As mentioned above, no available knowledge can be obtained when $t$=1. At $t$=3, both two algorithms use first-order temporal smoothness. At time step 5, the second-order knowledge is used, and $w$={0.9, 0.1} because the overlapping ratio $r_{35}$ is greater than $r_{15}$. At time step 7, the second-order knowledge is also used for simplicity, and $w$={0.7, 0.3} due to the relatively high similarity between $r_{57}$ and $r_{37}$. We can obtain the same claim at time step 9. When $t$=3, 5, 7, and 9, significant evolution is detected, and HoKT utilizes higher-order knowledge to obtain the maximum value of NMI. The performance of HoKT is improved by about 0.4% compared to DYNMOGA.

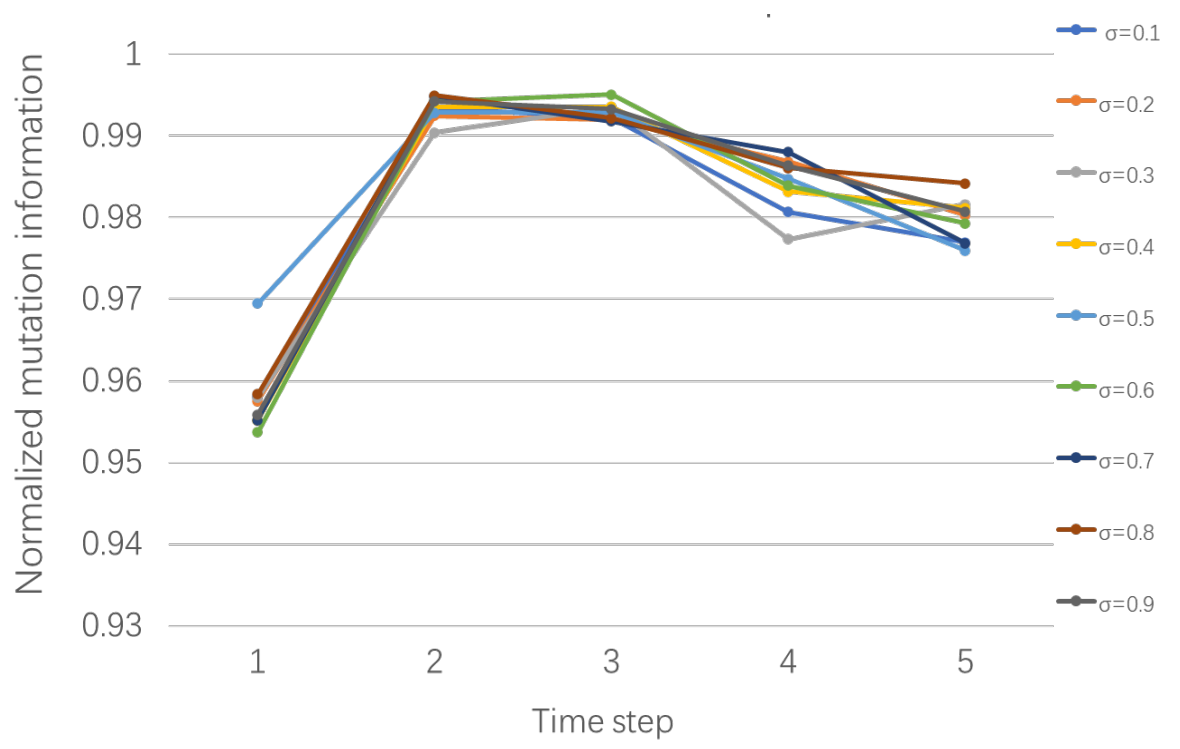

Figure 15 The change of NMI for different thresholds.

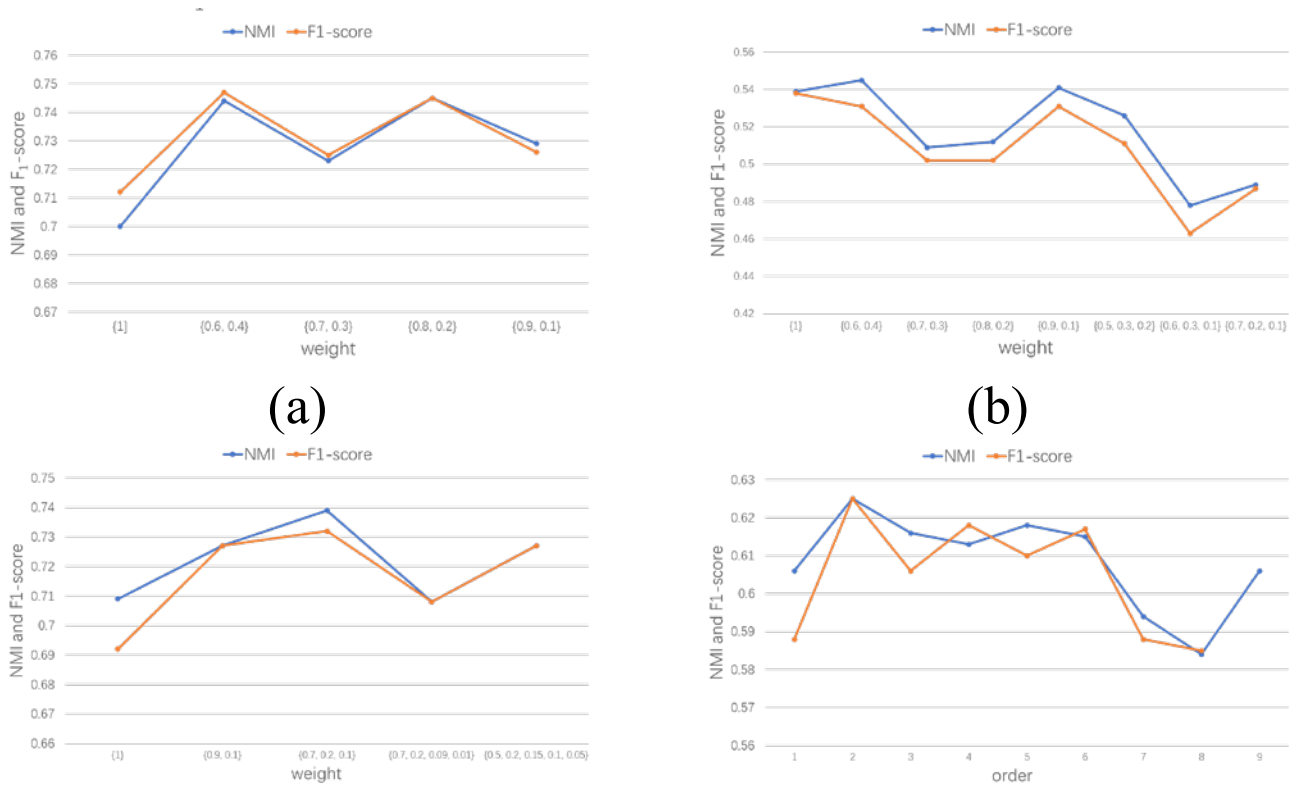

(a) (b) (c) (d)

Figure 16 NMI and $F_1$-score with different orders at (a) $t$=3, (b) $t$=4, (c) $t$=6, and (d) $t$=10.

## *E. Four Events*

The fifth dataset consists of four kinds of main events. There are five-time steps in each snapshot. The descriptions of these datasets are shown as follows:

1) Birth and Death: 10% of communities are restructured by selecting nodes from other existing communities and removing nodes from origin clusters.
2) Expansion and Contraction: 10% of communities are expanded or reduced by 25% of the original scale.
3) Intermittent Communities: 10% of communities are concealed and present in the next step.
4) Merging and Splitting: parts of the different communities are selected to merge at each time step.

Four datasets are generated based on four kinds of events. Each network consists of 1000 nodes whose mean degree is 15. In order to compare HoKT with other algorithms in the network with great changes, we run algorithms only at time steps 1, 3, and 5. Figs. 11-14 show the similarity matrix for the Four Events dataset. Table VIII depicts the metric NMI on the four datasets using HoKT, DYNMOGA, and KT-MOEA/D. Table IX shows the values of the $F_1$-score for HoKT, DYNMOGA, and KT-MOEA/D.

When $t$=1, HoKT uses the static community detection method by optimizing (2), and we select the worst NMI and $F_1$-score values to show the effectiveness of HoKT. Thus, it is reasonable that the performance of HoKT is worse than DYNMOGA and KT-MOEA/D. When $t$=3, the first-order knowledge is used. When $t$ = 5, the first-order or the second-order knowledge is used, and $w$ is listed in the table. NMI and $F_1$-score at time step 5 are mainly focused because the higher-order knowledge is used only in time step 5. Table VIII shows that HoKT outperforms or matches DYNMOGA in 11 out of 12 cases, only losing once to DYNMOGA at the first step. Furthermore, the $F_1$-score of HoKT is higher than those computed on the second and fourth networks. In the case of expansion and contraction communities, the NMI of DYNMOGA and KT-MOEA/D only decreased by about 0.01 and 0.04, respectively, from time step 3 to time step 5. Since the similarities between neighboring snapshots are more significant than 0.95, there is much helpful knowledge on first-order knowledge. However, the NMI of HoKT remains increased over the third time step. As regards the fourth dataset, although DYNMOGA and KT-MOEA/D obtained higher NMI values than HoKT in the first and second-time steps, NMI values obtained by DYNMOGA and KT-MOEA/D at time step 5 are lower than that obtained by HoKT. Moreover, the extra knee-based transfer strategy in KT-MOEA/D results in high performance in the Intermittent Communities dataset. For the other three datasets, HoKT is better than KT-MOEA/D. Thus, higher-order knowledge transfer is effective.

## *F. Parameters Analysis*

### *1)* ***Effect of σ on HoKT***

In this paper, we determine $\sigma$ by operating trial-and-error experiments on datasets. We set $\sigma$ = {0.2, 0.3, 0.4, 0.5, 0.6, 0.7, 0.8, 0.9} to run the algorithm on the benchmark merging and splitting dataset. The results are shown in Fig. 15. The red curve in Fig. 8 has the best NMI at each time step. The red data occur when $\sigma$ = 0.7, 0.8, and 0.9. That is because the similarities of the network are around [0.7, 0.9]. Furthermore, only when $\sigma$=0.7, 0.8, or 0.9, different order knowledge can be used. Furthermore, there are two red data when $\sigma$ = 0.8. Thus, we set $\sigma$ = 0.8. Moreover, we mainly migrate the second- and third-order knowledge and give weight to the different knowledge. We choose the best solution with different weights and orders as the final solution of HoKT.

It is concluded that the selection of $\sigma$ is based on the network similarity. At first, the similarity is computed to determine the value interval of $\sigma$. And then, the value of $\sigma$ is obtained by operating trial-and-error experiments. If more than two snapshots have a high overlapping ratio with the current snapshot, we can use higher-order knowledge; otherwise, we use first-order knowledge. It should be noted that if the similarity of two adjacent snapshots is low, we should use zero-order knowledge, i.e., not use any previous results.

TABLE X
NMI AND $F_1$-SCORE WITH DIFFERENT ORDERS WHEN T = 3.

| order | $w$ | NMI | $F_1$-score |
|---|---|---|---|
| 1 | {1} | 0.700±0.04 | 0.712±0.04 |
| 2 | {0.6, 0.4} | **0.744±0.05** | **0.747±0.04** |
| | {0.7, 0.3} | 0.723±0.05 | 0.725±0.05 |
| | {0.8, 0.2} | **0.745±0.05** | **0.745±0.05** |
| | {0.9, 0.1} | 0.729±0.04 | 0.726±0.04 |

TABLE XI
NMI AND $F_1$-SCORE WITH DIFFERENT ORDERS WHEN $T$ = 4.

| order | $w$ | NMI | $F_1$-score |
|---|---|---|---|
| 1 | {1} | 0.539±0.07 | 0.538±0.07 |
| 2 | {0.6, 0.4} | 0.545±0.05 | 0.531±0.06 |
| | {0.7, 0.3} | 0.509±0.10 | 0.502±0.10 |
| | {0.8, 0.2} | 0.512±0.07 | 0.502±0.08 |
| | {0.9, 0.1} | 0.541±0.07 | 0.531±0.10 |
| 3 | {0.5, 0.3, 0.2} | 0.526±0.01 | 0.511±0.01 |
| | {0.6, 0.3, 0.1} | 0.478±0.09 | 0.463±0.05 |
| | {0.7, 0.2, 0.1} | 0.489±0.00 | 0.487±0.01 |
| | {0.2, 0.3, 0.5} | **0.551±0.03** | **0.543±0.04** |

TABLE XII
NMI AND $F_1$-SCORE WITH DIFFERENT ORDERS WHEN $T$ = 6.

| order | $w$ | NMI | $F_1$-score |
|---|---|---|---|
| 1 | {1} | 0.709±0.06 | 0.692±0.06 |
| 2 | {0.6, 0.4} | 0.703±0.06 | 0.690±0.08 |
| | {0.7, 0.3} | 0.713±0.06 | 0.702±0.08 |
| | {0.8, 0.2} | 0.705±0.05 | 0.699±0.05 |
| | {0.9, 0.1} | 0.727±0.04 | 0.727±0.06 |
| 3 | {0.5, 0.3, 0.2} | 0.691±0.00 | 0.689±0.00 |
| | {0.6, 0.3, 0.1} | **0.737±0.07** | **0.732±0.07** |
| | {0.7, 0.2, 0.1} | **0.739±0.00** | **0.732±0.00** |
| 4 | {0.4, 0.3, 0.2, 0.1} | 0.687±0.05 | 0.657±0.10 |
| | {0.5, 0.3, 0.15, 0.05} | 0.704±0.05 | 0.704±0.05 |
| | {0.6, 0.2, 0.15, 0.05} | 0.682±0.05 | 0.667±0.07 |
| | {0.7, 0.2, 0.09, 0.01} | 0.708±0.04 | 0.708±0.04 |
| 5 | {0.5, 0.2, 0.15, 0.1, 0.05} | 0.727±0.07 | 0.727±0.06 |
| | {0.6, 0.2, 0.1, 0.09, 0.01} | 0.718±0.05 | 0.719±0.05 |

TABLE XIII
NMI AND $F_1$-SCORE WITH DIFFERENT ORDERS WHEN $T$ = 10.

| order | $w$ | NMI | $F_1$-score |
|---|---|---|---|
| 1 | {1} | 0.606±0.03 | 0.588±0.06 |
| 2 | {0.5, 0.5} | **0.627±0.06** | **0.626±0.05** |
| | {0.6, 0.4} | 0.593±0.06 | 0.594±0.06 |
| | {0.7, 0.3} | **0.625±0.05** | **0.625±0.05** |
| | {0.8, 0.2} | 0.606±0.05 | 0.607±0.05 |
| | {0.9, 0.1} | 0.593±0.06 | 0.584±0.07 |
| 3 | {0.5, 0.3, 0.2} | 0.606±0.00 | 0.605±0.00 |
| | {0.6, 0.3, 0.1} | 0.616±0.07 | 0.606±0.05 |
| | {0.7, 0.2, 0.1} | 0.600±0.00 | 0.603±0.00 |
| 4 | {0.5, 0.3, 0.15, 0.05} | 0.613±0.05 | 0.612±0.05 |
| | {0.6, 0.2, 0.15, 0.05} | 0.586±0.04 | 0.585±0.04 |
| | {0.7, 0.2, 0.09, 0.01} | 0.610±0.05 | 0.618±0.05 |
| 5 | {0.5, 0.2, 0.15, 0.1, 0.05} | 0.618±0.03 | 0.610±0.05 |
| | {0.6, 0.2, 0.1, 0.09, 0.01} | 0.606±0.04 | 0.605±0.04 |
| 6 | {0.4, 0.3, 0.15, 0.1, 0.03, 0.02} | 0.615±0.05 | 0.617±0.05 |
| 7 | {0.4, 0.3, 0.1, 0.08, 0.06, 0.04, 0.02 } | 0.594±0.07 | 0.588±0.08 |
| 8 | {0.4, 0.2, 0.1, 0.07, 0.05, 0.04, 0.03, 0.01} | 0.584±0.05 | 0.585±0.06 |
| 9 | {0.3, 0.2, 0.15, 0.09, 0.08, 0.07, 0.05, 0.04, 0.02} | 0.606±0.04 | 0.611±0.04 |

### *2) Effect of w on HoKT*

We study the influence of order and different weights on HoKT in terms of NMI and $F_1$-score. Thus, we run HoKT with different orders to select the most suitable one and take the Cell Phone Call dataset as an example. We choose several typical snapshots to analyze the impact of different weight combinations, including $t$=3, 4, 6, and 10.

We can find the great impact of different weight combinations on the algorithm's performance. We illustrate the weight selection strategy based on these cases. At time step 3, second-order knowledge can be used. The results are shown in Table X and Fig. 16(a). The NMI and $F_1$-score found by second-order knowledge are higher than those obtained using first-order knowledge. This case means that HoKT using higher-order knowledge provides a more structured division. How to determine the weight? The overlapping ratio $r_{13}$ of snapshots between $t$=1 and $t$=3 is 0.7506 and $r_{23}$=0.7600. The difference between those two $r$ values is noticeable, and $r_{23}$ is greater than $r_{13}$. Thus, the weight for $t$=2 should be greater than that for $t$=1. We test four combinations for $w$ and find that $w$={0.6, 0.4} and $w$={0.8, 0.2} are better than others.

The experimental results of different orders at time step 4 are shown in Table XI and Fig. 16(b). The overlapping ratio $r_{34}$ of snapshots between $t$=4 and $t$=3 is 0.7555, $r_{24}$=0.7540, and $r_{14}$=0.7568. The difference between them is slight and only 0.0015. Interestingly, $r_{14}$ has the highest proportion, so the proportion of weights should be the highest for the first snapshot and the lowest for the second snapshot. If the second-order knowledge is used, the weight for $t$=3 should be greater than that for $t$=2. $w$={0.6, 0.4} can obtain the best accuracy at time step 4. $w$={0.2, 0.3, 0.5} can be selected if the second-order knowledge is employed.

The experimental results of different orders at time step 6 are shown in Table XII and Fig. 16(c). The network at $t$=6 is more similar to that at $t$=5 and 4 than at $t$=3, 2, and 1. Thus, second-order knowledge or second-order knowledge is only used. The weight for the fifth snapshot is greater than the fourth and third snapshots. We find HoKT achieves the best performance with $w$={0.7, 0.2, 0.1} or $w$={0.6, 0.3, 0.1}. We also find that third-order knowledge is better than second-order knowledge. Although the weight of the third snapshot is only 0.1, it can bring a big improvement.

The experimental results of different orders at time step 10 are shown in Table XIII and Fig. 16(d). At time step 10, $r_{9,10}$, $r_{8,10}$, and $r_{7,10}$ rank top three, and $r_{9,10}$=$r_{8,10}$. Thus, if we select second-order knowledge and $w$={0.5, 0.5} whose performance is similar to that of $w$={0.7, 0.3}. If we select third-order knowledge and $w$={0.6, 0.3, 0.1}. The great overlapping ratio illustrates more useful knowledge in this snapshot for the next snapshot. Although the results with higher-order knowledge are not optimal, the bias is small, and a relatively satisfactory solution can be obtained, which may be attributed to the ensemble idea in the proposed scheme. Although we obtain the best performance with second-order knowledge, the performance gap between different combinations is large (large bias). Therefore, for new datasets, it is recommended to select slightly higher-order knowledge.

## V. CONCLUSION

HoKT has first gravitated to dynamic community detection problems over networks with great changes, which is essential in many fields. The proposed higher-order knowledge transfer strategy with *HONMI* can balance the positive and negative transfer caused by great changes among snapshots. HoKT adapts to various scenarios, including network sizes, similarity distributions, and minor or great changes between snapshots. In the case of great changes, the performance improvement of HoKT is noticeable, and we still have a good performance in the case of minor changes. Moreover, if the overlapping rate is high (>0.95), the gain brought by higher-order knowledge is not apparent but positive because the first-order knowledge already contains rich information.

TABLE AI
NMI FOR FOUR DATASETS.

| Datasets | time | HoKT | DYNMOGA | KT-MOEA/D |
|---|---|---|---|---|
| 1 | $t = 1$ | **0.9684±0.0141** | **0.9560±0.0198** | **0.9649±0.0074** |
| | $t = 2$ | **0.9988±0.0021** | **0.9968±0.0027** | 0.9957±0.0033 |
| | $t = 3$ | **0.9985±0.0022** | **0.9984±0.0029** | **0.9977±0.0030** |
| | $t = 4$ | **0.9997±0.0001** | **0.9976±0.0018** | **0.9977±0.0014** |
| | $t = 5$ | **0.9989±0.0013** | **0.9970±0.0021** | **0.9992±0.0011** |
| 2 | $t = 1$ | **0.9588±0.0148** | **0.9551±0.0179** | 0.9421±0.0151 |
| | $t = 2$ | **0.9957±0.0036** | **0.9939±0.0042** | 0.9773±0.0072 |
| | $t = 3$ | **0.9954±0.0034** | 0.9922±0.0058 | 0.9734±0.0087 |
| | $t = 4$ | **0.9890±0.0113** | 0.9846±0.0114 | 0.9509±0.0177 |
| | $t = 5$ | **0.9868±0.0057** | 0.9815±0.0078 | 0.9335±0.0159 |
| 3 | $t = 1$ | **0.9612±0.0098** | **0.9649±0.0103** | 0.9354±0.0189 |
| | $t = 2$ | **0.9957±0.0031** | **0.9963±0.0018** | 0.9819±0.0054 |
| | $t = 3$ | **0.9951±0.0028** | **0.9965±0.0031** | 0.9812±0.0081 |
| | $t = 4$ | **0.9953±0.0036** | **0.9961±0.0024** | 0.9805±0.0085 |
| | $t = 5$ | **0.9933±0.0062** | **0.9945±0.0036** | 0.9765±0.0071 |
| 4 | $t = 1$ | **0.9629±0.0025** | **0.9649±0.0117** | 0.9331±0.0236 |
| | $t = 2$ | **0.9993±0.0005** | 0.9979±0.0020 | 0.9851±0.0076 |
| | $t = 3$ | **0.8678±0.0010** | **0.8660±0.0022** | 0.8522±0.0054 |
| | $t = 4$ | **0.7626±0.0022** | **0.7623±0.0009** | 0.7558±0.0039 |
| | $t = 5$ | **0.6972±0.0012** | **0.6972±0.0013** | 0.6905±0.0072 |
| win/tie/loss | | - | 0/16/4 | 0/4/16 |

TABLE AII
$F_1$-SCORE FOR FOUR DATASETS.

| Datasets | Time | HoKT | DYNMOGA | KT-MOEA/D |
|---|---|---|---|---|
| 1 | $t = 1$ | **0.9684±0.0069** | **0.9615±0.0117** | **0.9645±0.0081** |
| | $t = 2$ | **0.9989±0.0008** | **0.9970±0.0038** | 0.9954±0.0029 |
| | $t = 3$ | **0.9986±0.0010** | **0.9981±0.0008** | 0.9970±0.0027 |
| | $t = 4$ | **0.9997±0.0001** | 0.9984±0.0011 | 0.9982±0.0013 |
| | $t = 5$ | **0.9999±0.0001** | 0.9979±0.0008 | **0.9992±0.0007** |
| 2 | $t = 1$ | **0.9593±0.0125** | **0.9551±0.0158** | **0.9532±0.0174** |
| | $t = 2$ | **0.9984±0.0015** | **0.9939±0.0043** | **0.9934±0.0041** |
| | $t = 3$ | **0.9972±0.0015** | 0.9922±0.0057 | **0.9928±0.0081** |
| | $t = 4$ | **0.9983±0.0018** | 0.9841±0.0128 | 0.9893±0.0054 |
| | $t = 5$ | **0.9982±0.0013** | 0.9811±0.0070 | 0.9825±0.0064 |
| 3 | $t = 1$ | **0.9646±0.0094** | 0.9551±0.0106 | **0.9628±0.0103** |
| | $t = 2$ | **0.9978±0.0019** | **0.9986±0.0017** | **0.9957±0.0034** |
| | $t = 3$ | **0.9985±0.0017** | **0.9985±0.0012** | 0.9967±0.0031 |
| | $t = 4$ | **0.9981±0.0002** | 0.9930±0.0043 | 0.9957±0.0021 |
| | $t = 5$ | **0.9950±0.0046** | **0.9927±0.0029** | **0.9940±0.0021** |
| 4 | $t = 1$ | **0.9629±0.0096** | **0.9527±0.0176** | **0.9602±0.0202** |
| | $t = 2$ | **0.9909±0.0016** | **0.9928±0.0011** | **0.9897±0.0019** |
| | $t = 3$ | **0.8511±0.0009** | **0.8508±0.0014** | **0.8499±0.0014** |
| | $t = 4$ | **0.7452±0.0020** | **0.7446±0.0024** | **0.7454±0.0020** |
| | $t = 5$ | **0.6545±0.0011** | **0.6546±0.0021** | **0.6538±0.0017** |
| win/tie/loss | | - | 0/13/7 | 0/13/7 |

The concept of high-order knowledge transfer can be extended to other dynamic optimization problems with great changes, not limited to the current community detection problem, providing a new solution. Also, the parameter $w$ needs to be set by users in our algorithm. In the future, we can design a self-adaptive algorithm to obtain a better $w$.

## APPENDIX

Four datasets are generated based on the four different kinds of events. Each network consists of 1000 nodes whose mean degree is 15. Table AI depicts NMI on four networks in terms of three different algorithms. Table AII shows $F_1$-score in terms of HoKT, DYNMOGA, and KT-MOEA/D. When $t = 2$, the first-order knowledge is used. When $t = 3$, the second-order knowledge is used and $w = \{0.8, 0.2\}$. When $t = 4$, the third-order knowledge is used and $w = \{0.7, 0.2, 0.1\}$. When $t = 5$, the second-order knowledge is used and $w = \{0.7, 0.3\}$. The gain brought by higher-order knowledge is not obvious but positive because the first-order knowledge already contains rich information. The running time of HoKT for all datasets is shown in Fig. A1. As the size of the network continues to increase, the algorithm's runtime becomes longer.

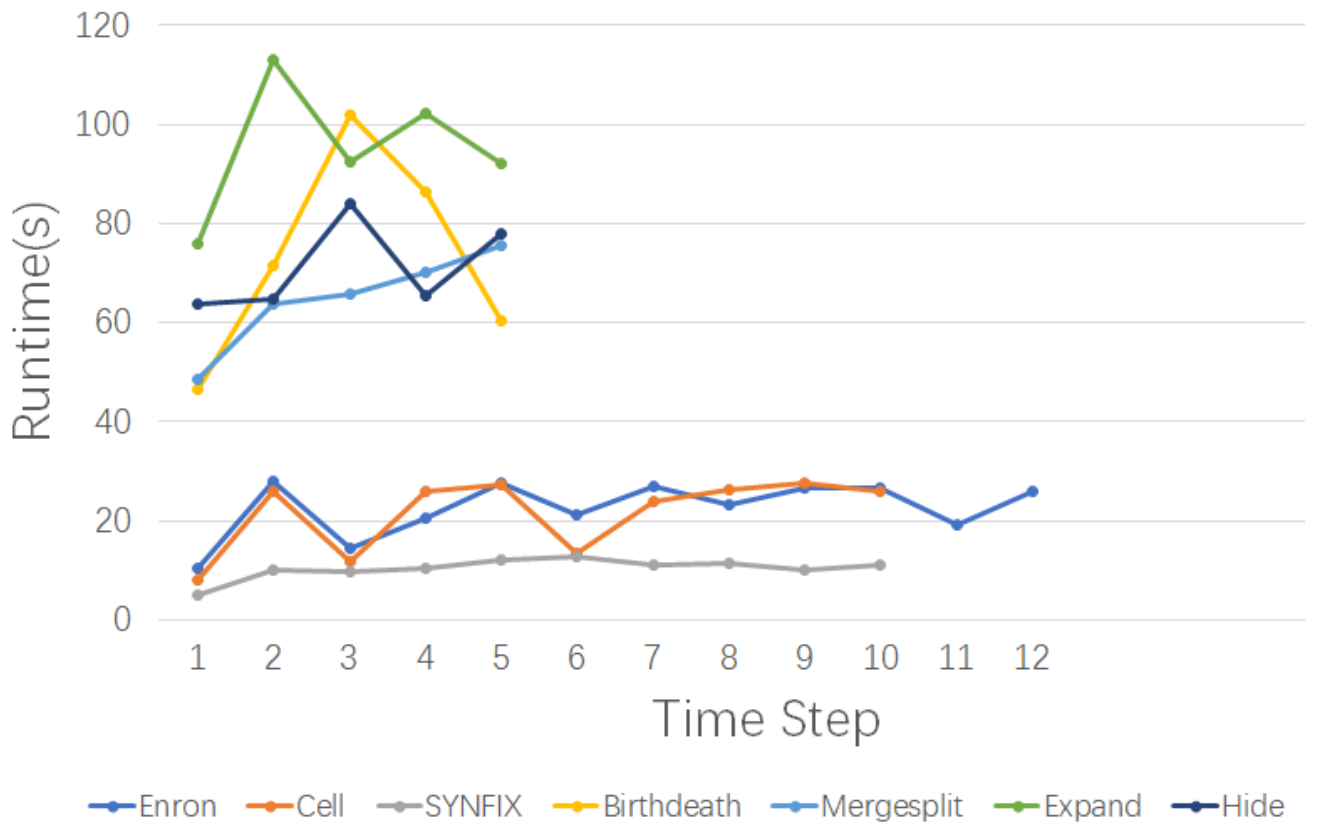

Figure A1 The running time of HoKT for all datasets.

# Supplementary Material of "Higher-order Knowledge Transfer for Dynamic Community Detection with Great Changes"

Huixin Ma, Kai Wu, *Member*, *IEEE*, Handing Wang, *Member*, *IEEE*, Jing Liu, *Senior Member*, *IEEE*

***Abstract*—This section is the supplementary material for HoKT and mainly contains the basic properties of the used dataset and additional results of the parametric analysis section.**

## I. Parameter Analysis

We study the influence of order and different weights on HoKT in terms of NMI and $F_1$-score. Thus, we run HoKT with different orders to select the most suitable one and take the Cell Phone Call dataset as an example. In the first time step, the static method is used. At time step 2, there is only first-order knowledge. In time step 3, we use first-order knowledge and second-order knowledge. The results are shown in Table I and Fig. 1(a). As shown in Fig. 1(a), both the NMI and $F_1$-score found by second-order knowledge are higher than those obtained using first-order knowledge. This means that HoKT using higher-order knowledge provides a more structured division. The high similarity of the second-order network shows positive knowledge in second-order knowledge. Thus, when the second-order knowledge is selected and $w$ = {0.6, 0.4}, HoKT gets the best NMI.

TABLE I
NMI and $F_1$-score with different orders when T = 3.

| order | $w$ | NMI | $F_1$-score |
|---|---|---|---|
| 1 | {1} | 0.700±0.04 | 0.712±0.04 |
| 2 | {0.6, 0.4} | **0.744±0.05** | **0.747±0.04** |
| | {0.7, 0.3} | 0.723±0.05 | 0.725±0.05 |
| | {0.8, 0.2} | **0.745±0.05** | **0.745±0.05** |
| | {0.9, 0.1} | 0.729±0.04 | 0.726±0.04 |

TABLE II
NMI and $F_1$-score with different orders when $T$ = 4.

| order | $w$ | NMI | $F_1$-score |
|---|---|---|---|
| 1 | {1} | 0.539±0.07 | 0.538±0.07 |
| 2 | {0.6, 0.4} | 0.545±0.05 | 0.531±0.06 |
| | {0.7, 0.3} | 0.509±0.10 | 0.502±0.10 |
| | {0.8, 0.2} | 0.512±0.07 | 0.502±0.08 |
| | {0.9, 0.1} | 0.541±0.07 | 0.531±0.10 |
| 3 | {0.5, 0.3, 0.2} | 0.526±0.01 | 0.511±0.01 |
| | {0.6, 0.3, 0.1} | 0.478±0.09 | 0.463±0.05 |
| | {0.7, 0.2, 0.1} | 0.489±0.00 | 0.487±0.01 |
| | **{0.2, 0.3, 0.5}** | **0.551±0.03** | **0.543±0.04** |

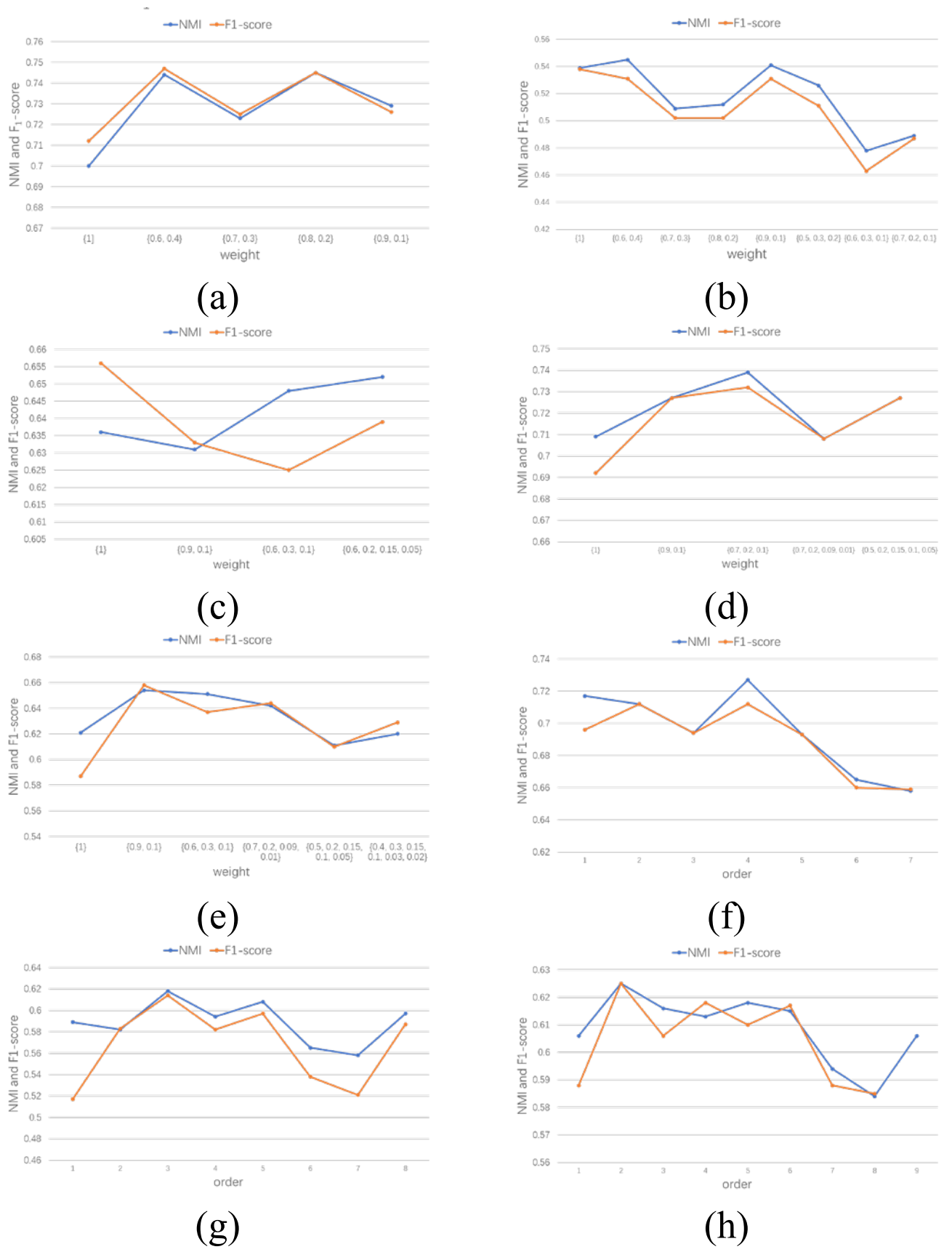

Figure 1 NMI and $F_1$-score with different orders at (a) $t$ = 3, (b) $t$ = 4, (c) $t$ = 5, (d), $t$ = 6, (e) $t$ = 7, (f) $t$ = 8, (g) $t$ = 9, and (h) $t$ = 10.

1) The order is set as 1, 2, 3, and 4 to run HoKT. The experimental results of different orders at time step 5 are shown in Table III and Fig. 1(b). In Fig. 2, networks at time steps 1, 2, and 3 are more similar to the network at time step 5 than the network at time step 4. The higher-order knowledge can balance the negative knowledge in first-order knowledge. The higher the order is, the greater the NMI we can obtain. It shows that when the network change occurs, the higher-order

This work was supported by the Key Project of Science and Technology Innovation 2030 supported by the Ministry of Science and Technology of China under Grant 2018AAA0101302.

Huixin Ma, Kai Wu, and Handing Wang are with the School of Artificial Intelligence, Xidian University, Xi'an 710071, China (e-mail: 948181384@qq.com kwu@xidian.edu.cn, hdwang@xidian.edu.cn).

Jing Liu is with the Guangzhou Institute of Technology, Xidian University, Guangzhou 510555, China (e-mail: neouma@mail.xidian.edu.cn).

TABLE III
NMI AND $F_1$-SCORE WITH DIFFERENT ORDERS WHEN $T = 5$.

| order | w | NMI | $F_1$-score |
|---|---|---|---|
| 1 | {1} | 0.636±0.08 | 0.656±0.09 |
| 2 | {0.6, 0.4} | 0.626±0.05 | 0.607±0.07 |
| | {0.7, 0.3} | 0.624±0.09 | 0.620±0.09 |
| | {0.8, 0.2} | 0.615±0.07 | 0.588±0.09 |
| | {0.9, 0.1} | 0.631±0.04 | 0.633±0.05 |
| 3 | {0.5, 0.3, 0.2} | 0.628±0.00 | 0.613±0.01 |
| | {0.6, 0.3, 0.1} | 0.648±0.05 | 0.625±0.09 |
| | {0.7, 0.2, 0.1} | 0.622±0.00 | 0.599±0.01 |
| 4 | {0.4, 0.3, 0.2, 0.1} | 0.607±0.08 | 0.608±0.08 |
| | {0.5, 0.3, 0.15, 0.05} | 0.632±0.08 | 0.601±0.10 |
| | {0.6, 0.2, 0.15, 0.05} | **0.652±0.07** | **0.639±0.08** |
| | {0.7, 0.2, 0.09, 0.01} | 0.649±0.05 | 0.638±0.06 |

TABLE IV
NMI AND $F_1$-SCORE WITH DIFFERENT ORDERS WHEN $T = 6$.

| order | w | NMI | $F_1$-score |
|---|---|---|---|
| 1 | {1} | 0.709±0.06 | 0.692±0.06 |
| 2 | {0.6, 0.4} | 0.703±0.06 | 0.690±0.08 |
| | {0.7, 0.3} | 0.713±0.06 | 0.702±0.08 |
| | {0.8, 0.2} | 0.705±0.05 | 0.699±0.05 |
| | {0.9, 0.1} | 0.727±0.04 | 0.727±0.06 |
| 3 | {0.5, 0.3, 0.2} | 0.691±0.00 | 0.689±0.00 |
| | {0.6, 0.3, 0.1} | **0.737±0.07** | **0.732±0.07** |
| | {0.7, 0.2, 0.1} | **0.739±0.00** | **0.732±0.00** |
| 4 | {0.4, 0.3, 0.2, 0.1} | 0.687±0.05 | 0.657±0.10 |
| | {0.5, 0.3, 0.15, 0.05} | 0.704±0.05 | 0.704±0.05 |
| | {0.6, 0.2, 0.15, 0.05} | 0.682±0.05 | 0.667±0.07 |
| | {0.7, 0.2, 0.09, 0.01} | 0.708±0.04 | 0.708±0.04 |
| 5 | {0.5, 0.2, 0.15, 0.1, 0.05} | 0.727±0.07 | 0.727±0.06 |
| | {0.6, 0.2, 0.1, 0.09, 0.01} | 0.718±0.05 | 0.719±0.05 |

TABLE V
NMI AND $F_1$-SCORE WITH DIFFERENT ORDERS WHEN $T = 7$.

| order | w | NMI | $F_1$-score |
|---|---|---|---|
| 1 | {1} | 0.621±0.06 | 0.587±0.11 |
| 2 | {0.6, 0.4} | 0.623±0.06 | 0.613±0.06 |
| | {0.7, 0.3} | 0.646±0.06 | 0.646±0.06 |
| | {0.8, 0.2} | 0.645±0.09 | 0.622±0.11 |
| | {0.9, 0.1} | **0.654±0.04** | **0.658±0.06** |
| 3 | {0.5, 0.3, 0.2} | 0.584±0.00 | 0.562±0.01 |
| | {0.6, 0.3, 0.1} | 0.651±0.05 | 0.637±0.07 |
| | {0.7, 0.2, 0.1} | 0.606±0.00 | 0.604±0.00 |
| 4 | {0.4, 0.3, 0.2, 0.1} | 0.602±0.07 | 0.575±0.09 |
| | {0.5, 0.3, 0.15, 0.05} | 0.616±0.07 | 0.584±0.09 |
| | {0.6, 0.2, 0.15, 0.05} | 0.606±0.08 | 0.611±0.11 |
| | {0.7, 0.2, 0.09, 0.01} | 0.642±0.07 | 0.644±0.07 |
| 5 | {0.5, 0.2, 0.15, 0.1, 0.05} | 0.611±0.07 | 0.610±0.09 |
| | {0.6, 0.2, 0.1, 0.09, 0.01} | 0.611±0.06 | 0.596±0.08 |
| 6 | {0.4, 0.3, 0.15, 0.1, 0.03, 0.02} | 0.620±0.06 | 0.629±0.06 |

knowledge can adapt the algorithm to the new environment faster and eventually improve the precision of the solution.

2) Table III and Fig. 1(c) display that HoKT runs using six orders at $t = 7$. We can find that HoKT still achieves better solutions when higher-order knowledge is used compared with using first-order knowledge. It is tough to find the similarity between the network at $t = 7$ and the network at $t = 1$ and 2, so the result when orders are 5 and 6 is not desirable. It is obvious that the network at $t = 6$ is more similar to other networks in Fig. 2. The second-order knowledge has more useful information because the great change occurs in two adjacent periods, and second-order knowledge is still advisable.

3) The NMI and $F_1$-score with different orders are reported in Table VII and Fig. 1(d). The results when the order is set to 3 are notably better than those with other orders. In particular, the NMI values produced by the method using third-order knowledge cover 5% higher than the algorithm using first-order knowledge. In addition, higher NMI scores and $F_1$-score of HoKT imply that network communities identified by our approach are not only highly similar to the ground truth but also more precise than that detected by DYNMOGA.

TABLE VI
NMI AND $F_1$-SCORE WITH DIFFERENT ORDERS WHEN $T = 8$.

| order | w | NMI | $F_1$-score |
|---|---|---|---|
| 1 | {1} | 0.717±0.06 | 0.696±0.05 |
| 2 | {0.6, 0.4} | 0.675±0.09 | 0.673±0.09 |
| | {0.7, 0.3} | 0.712±0.06 | 0.712±0.06 |
| | {0.8, 0.2} | 0.671±0.09 | 0.671±0.09 |
| | {0.9, 0.1} | 0.693±0.06 | 0.691±0.06 |
| 3 | {0.5, 0.3, 0.2} | 0.671±0.00 | 0.672±0.00 |
| | {0.6, 0.3, 0.1} | 0.690±0.05 | 0.689±0.05 |
| | {0.7, 0.2, 0.1} | 0.694±0.00 | 0.694±0.00 |
| 4 | {0.5, 0.3, 0.15, 0.05} | 0.698±0.05 | 0.698±0.05 |
| | {0.6, 0.2, 0.15, 0.05} | 0.676±0.06 | 0.668±0.06 |
| | {0.7, 0.2, 0.09, 0.01} | **0.727±0.07** | **0.712±0.06** |
| 5 | {0.5, 0.2, 0.15, 0.1, 0.05} | 0.656±0.06 | 0.656±0.06 |
| | {0.6, 0.2, 0.1, 0.09, 0.01} | 0.693±0.03 | 0.693±0.03 |
| 6 | {0.4, 0.3, 0.15, 0.1, 0.03, 0.02} | 0.665±0.07 | 0.660±0.08 |
| 7 | {0.4, 0.3, 0.1, 0.08, 0.06, 0.04, 0.02} | 0.658±0.06 | 0.659±0.06 |

TABLE VII
NMI AND $F_1$-SCORE WITH DIFFERENT ORDERS WHEN $T = 9$.

| order | w | NMI | $F_1$-score |
|---|---|---|---|
| 1 | {1} | 0.589±0.07 | 0.517±0.09 |
| 2 | {0.6, 0.4} | 0.582±0.05 | 0.576±0.06 |
| | {0.7, 0.3} | 0.573±0.06 | 0.553±0.07 |
| | {0.8, 0.2} | 0.574±0.06 | 0.542±0.07 |
| | {0.9, 0.1} | 0.579±0.07 | 0.583±0.07 |
| 3 | {0.5, 0.3, 0.2} | 0.604±0.00 | 0.597±0.00 |
| | {0.6, 0.3, 0.1} | 0.571±0.04 | 0.563±0.04 |
| | {0.7, 0.2, 0.1} | **0.618±0.00** | **0.614±0.00** |
| 4 | {0.5, 0.3, 0.15, 0.05} | 0.581±0.06 | 0.573±0.06 |
| | {0.6, 0.2, 0.15, 0.05} | 0.581±0.11 | 0.582±0.11 |
| | {0.7, 0.2, 0.09, 0.01} | 0.594±0.06 | 0.578±0.07 |
| 5 | {0.5, 0.2, 0.15, 0.1, 0.05} | 0.608±0.06 | 0.597±0.07 |
| | {0.6, 0.2, 0.1, 0.09, 0.01} | 0.576±0.03 | 0.567±0.04 |
| 6 | {0.4, 0.3, 0.15, 0.1, 0.03, 0.02} | 0.565±0.08 | 0.538±0.11 |
| 7 | {0.4, 0.3, 0.1, 0.08, 0.06, 0.04, 0.02} | 0.558±0.06 | 0.521±0.09 |
| 8 | {0.4, 0.2, 0.1, 0.07, 0.05, 0.04, 0.03, 0.01} | 0.597±0.08 | 0.587±0.08 |

TABLE VIII
NMI AND $F_1$-SCORE WITH DIFFERENT ORDERS WHEN $T = 10$.

| order | w | NMI | $F_1$-score |
|---|---|---|---|
| 1 | {1} | 0.606±0.03 | 0.588±0.06 |
| 2 | {0.6, 0.4} | 0.593±0.06 | 0.594±0.06 |
| | {0.7, 0.3} | **0.625±0.05** | **0.625±0.05** |
| | {0.8, 0.2} | 0.606±0.05 | 0.607±0.05 |
| | {0.9, 0.1} | 0.593±0.06 | 0.584±0.07 |
| 3 | {0.5, 0.3, 0.2} | 0.606±0.00 | 0.605±0.00 |
| | {0.6, 0.3, 0.1} | 0.616±0.07 | 0.606±0.05 |
| | {0.7, 0.2, 0.1} | 0.600±0.00 | 0.603±0.00 |
| 4 | {0.5, 0.3, 0.15, 0.05} | 0.613±0.05 | 0.612±0.05 |
| | {0.6, 0.2, 0.15, 0.05} | 0.586±0.04 | 0.585±0.04 |
| | {0.7, 0.2, 0.09, 0.01} | 0.610±0.05 | 0.618±0.05 |
| 5 | {0.5, 0.2, 0.15, 0.1, 0.05} | 0.618±0.03 | 0.610±0.05 |
| | {0.6, 0.2, 0.1, 0.09, 0.01} | 0.606±0.04 | 0.605±0.04 |
| 6 | {0.4, 0.3, 0.15, 0.1, 0.03, 0.02} | 0.615±0.05 | 0.617±0.05 |
| 7 | {0.4, 0.3, 0.1, 0.08, 0.06, 0.04, 0.02 } | 0.594±0.07 | 0.588±0.08 |
| 8 | {0.4, 0.2, 0.1, 0.07, 0.05, 0.04, 0.03, 0.01} | 0.584±0.05 | 0.585±0.06 |
| 9 | {0.3, 0.2, 0.15, 0.09, 0.08, 0.07, 0.05, 0.04, 0.02} | 0.606±0.04 | 0.611±0.04 |

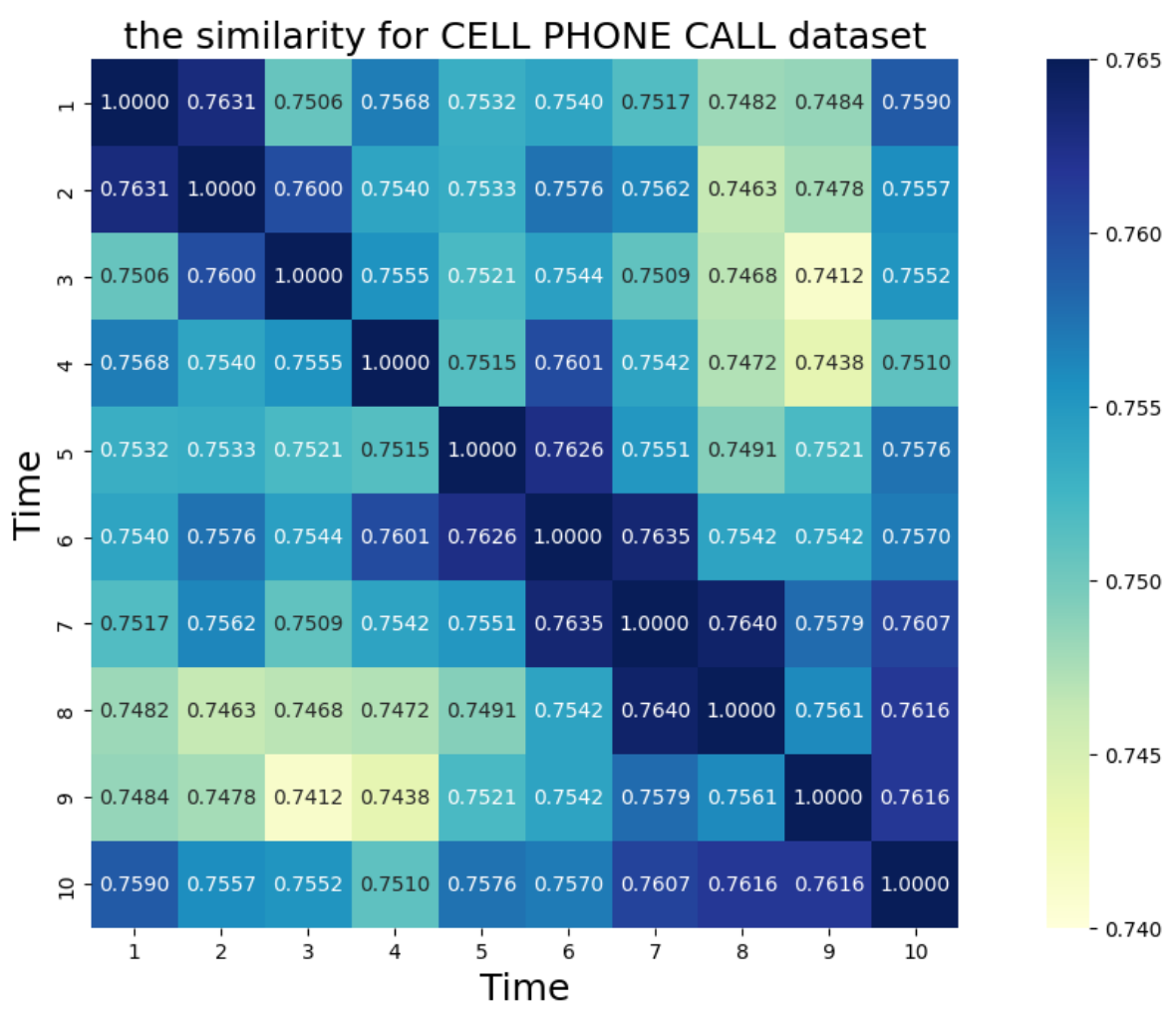

Figure 2 The similarity for CELL PHONE CALL dataset.

## II. Network Types

In this section, we describe the statistics of Cell and Enron Datasets, shown in Tables IX and X.

Four datasets used in this paper are common datasets used in dynamic community detection methods, including synthetic networks and real networks, and these representative datasets are provided with different properties. They can examine the applicability of our method in different situations.

(1) Different network types and sizes;

(2) The overlap rate between snapshots is distributed between [0.7, 1], which can examine the adaptability of our algorithm to problems with different overlap rates (see Main Text, Figs. 8-14);

(3) The overlap rate of some datasets continues to rise over time (Four Events Dataset), while the overlap rate of some datasets fluctuates continuously over time (Enron dataset). Through these datasets, we can examine the adaptability of HoKT in different higher-order knowledge scenarios;

(4) For example, in the Cell and Enron datasets (see Main Text, Tables IX and X), there are differences in nodes, edges, and communities between different snapshots. The generalization ability of our algorithm can be examined.

(5) Through manual settings, we consider the adaptability of our algorithm in the case of minor or great changes, respectively.

## III. Explanations of Two Objectives

**Why are the used objectives conflicting, i.e., modularity and NMI?** The reasons are shown as follows:

1) Current dynamic community detection methods often formulate these two objectives as a multi-objective problem.

2) The purpose of Modularity is to obtain the best community structure for a snapshot. NMI maximizes the similarity between the community structure at the last snapshot and the current community structure to ensure that the community label at the last snapshot can be used. However, the dynamic network characteristics significantly differ between the two snapshots. Therefore, the higher the NMI, the closer the community structure of the two snapshots is, and the optimal community structure of the current snapshot is destroyed, resulting in a decrease in the value of Modularity and vice versa. Thus, Modularity and NMI are conflicting. The Pareto front is shown as follows:

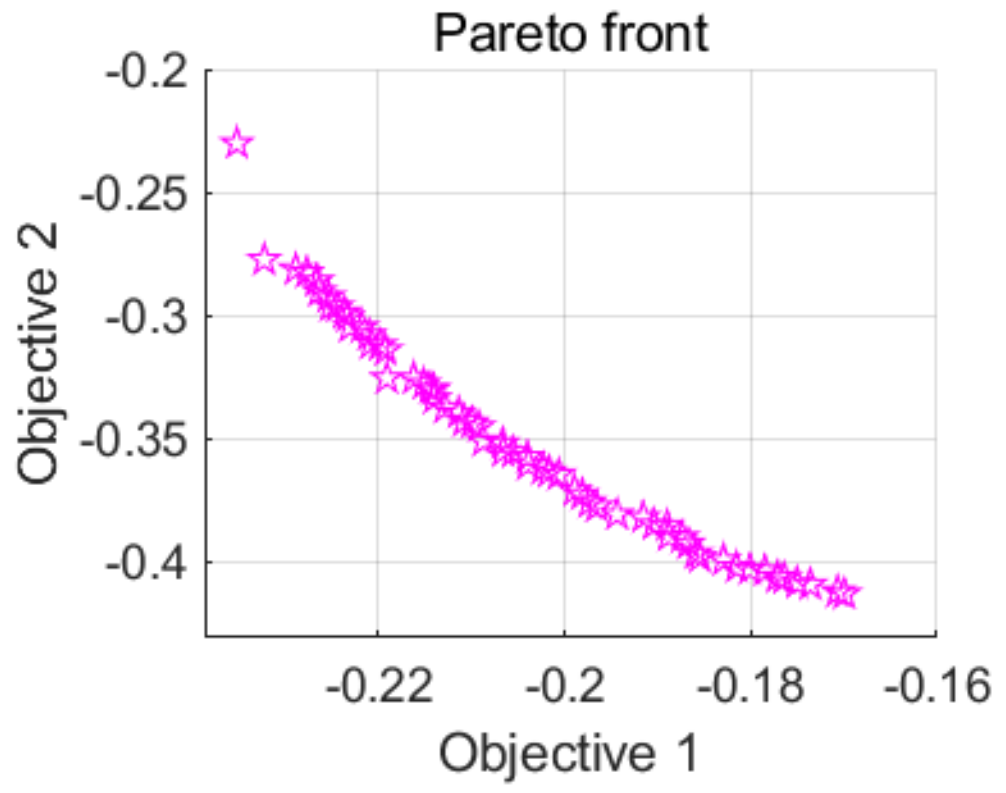

With the first objective function increasing, the second objective function decreases. Thus, modularity and NMI are conflicting.

TABLE IX
CELL DATASET STATISTICS. $T$ IS THE NUMBER OF TIME STEPS, $M$ THE MODULARITY, $|C|$ THE NUMBER OF COMMUNITIES, $|V|$ THE NUMBER OF NODES, $|E|$ THE NUMBER OF EDGES, $|E|^*$ THE NUMBER OF DISTINCT UNDIRECTED EDGES, AND $Z$ DENOTES THE MEAN DEGREE.

| $T$ | $M$ | $\|C\|$ | $\|V\|$ | $\|E\|$ | $\|E\|^*$ | $Z$ |
|---|---|---|---|---|---|---|
| 1 | 0.6640 | 32 | 370 | 987 | 525 | 2.6250 |
| 2 | 0.6561 | 35 | 373 | 964 | 499 | 2.4950 |
| 3 | 0.6587 | 30 | 374 | 953 | 509 | 2.5450 |
| 4 | 0.6540 | 31 | 374 | 1013 | 514 | 2.5700 |
| 5 | 0.6626 | 32 | 373 | 991 | 508 | 2.5400 |
| 6 | 0.6651 | 25 | 373 | 963 | 512 | 2.5600 |
| 7 | 0.6571 | 33 | 367 | 936 | 498 | 2.4900 |
| 8 | 0.6329 | 36 | 365 | 1005 | 511 | 2.5550 |
| 9 | 0.6538 | 34 | 374 | 982 | 518 | 2.5900 |
| 10 | 0.6467 | 32 | 384 | 1040 | 530 | 2.6500 |

TABLE X
ENRON DATASET STATISTICS.

| $T$ | $M$ | $\|C\|$ | $\|V\|$ | $\|E\|$ | $\|E\|^*$ | $Z$ |
|---|---|---|---|---|---|---|
| 1 | 0.6368 | 11 | 96 | 1070 | 180 | 2.3841 |
| 2 | 0.6803 | 7 | 93 | 1559 | 204 | 2.7020 |
| 3 | 0.6550 | 12 | 97 | 1844 | 218 | 2.8874 |
| 4 | 0.6571 | 12 | 108 | 1869 | 257 | 3.4040 |
| 5 | 0.5699 | 15 | 125 | 1919 | 292 | 3.8675 |
| 6 | 0.6943 | 10 | 120 | 1001 | 231 | 3.0596 |
| 7 | 0.6530 | 10 | 109 | 1325 | 252 | 3.3377 |
| 8 | 0.5356 | 9 | 131 | 2270 | 396 | 5.2450 |
| 9 | 0.6324 | 10 | 128 | 3152 | 361 | 4.7815 |
| 10 | 0.5302 | 13 | 135 | 8693 | 575 | 7.6159 |
| 11 | 0.5707 | 9 | 127 | 6276 | 469 | 6.2119 |
| 12 | 0.6152 | 8 | 113 | 2146 | 325 | 4.3046 |